\documentclass[11pt]{article}
\usepackage{acl}

\usepackage{times}
\usepackage{latexsym}
\usepackage[T1]{fontenc}
\usepackage[utf8]{inputenc}
\usepackage{microtype}
\usepackage{inconsolata}
\usepackage{graphicx}

\usepackage{amsmath}
\usepackage{amssymb}
\usepackage{booktabs}
\usepackage[dvipsnames,table]{xcolor}
\usepackage{colortbl}
\definecolor{bestcell}{RGB}{210,232,250}
\usepackage{amsthm}
\newtheorem{proposition}{Proposition}[section]
\newtheorem{lemma}[proposition]{Lemma}

\title{Forget Faster, Think Shorter: Efficient Reasoning from Recurrent Memory in Hybrid Language Models}

\author{
  Md\ Jueal Mia$^{1,2}$ and M.\ Hadi Amini$^{1,2}$ \\
  $^{1}$Knight Foundation School of Computing and Information Sciences \\
  $^{2}$Security, Optimization, and Learning for InterDependent networks laboratory (solid lab) \\
  Florida International University \\
  Miami, FL, USA \\
  \texttt{\{mmia001, moamini\}@fiu.edu}
}

\begin{document}
\maketitle
\begin{abstract}
Large reasoning models often keep thinking after they have reached an
answer. Existing methods retrain the model, learn steering directions from
curated data, or intervene during decoding. Hybrid language models offer a
different route: in their recurrent layers, how fast each head forgets is set
by a learned parameter stored in the checkpoint. We ask whether this decay
rate affects how long the model reasons. Multiplying every decay rate by one
factor $\kappa$ is a one-time checkpoint edit that requires no retraining or
decoding change; we selected $\kappa=2$ on a development set (MATH-Dev) and
kept it fixed for held-out evaluation. On Qwen3.5 Gated DeltaNet hybrids
(2B--9B parameters), reasoning length followed the decay rate in both
directions. At $\kappa=2$, thinking fell by 12.2--22.6\% on MATH-Dev, by
11.7--18.6\% on held-out MATH-500 and by 23.4--38.7\% on MATH-Vision
(testmini), with no detected loss of accuracy, while slower forgetting
($\kappa<1$) lengthened thinking in every text setting and lowered accuracy in
17 of 18. Accuracy gains, such as 66.5\% to 76.0\% for the 2B model on
MATH-Dev, came mainly from more traces finishing within the budget. In these
hybrids, memory decay is thus a simple, training-free control of reasoning
length.
\end{abstract}
\section{Introduction}
\label{sec:intro}

Large reasoning models solve hard problems by writing a long chain of
thought before they answer, and spending more of this test-time computation
has become one of the main ways to increase accuracy
\citep{deepseek2025r1,snell2024scaling,muennighoff2025s1}. This accuracy
comes at a cost. Reasoning tokens are produced one at a time, so the price
of an answer grows with the length of the reasoning behind it. Hybrid
language models reduce part of this cost through their architecture. They
combine attention with recurrent layers such as Gated DeltaNet
\citep{yang2025gdn} or Mamba-2 \citep{dao2024mamba2}, which compress the past
into a fixed-size state; most designs replace the majority of attention
layers with recurrent ones and keep a few attention layers for exact
look-up. Recent open reasoning models such as Qwen3.5, Nemotron Nano 2 and
Falcon-H1R are built on hybrid architectures
\citep{qwen35,nvidia2025nemotronnano2,falcon2026h1r}. In their recurrent
layers, how quickly the memory decays is controlled by a learned per-head
rate that multiplies an input-dependent step.

\begin{figure*}[t]
  \centering
  \includegraphics[width=\textwidth]{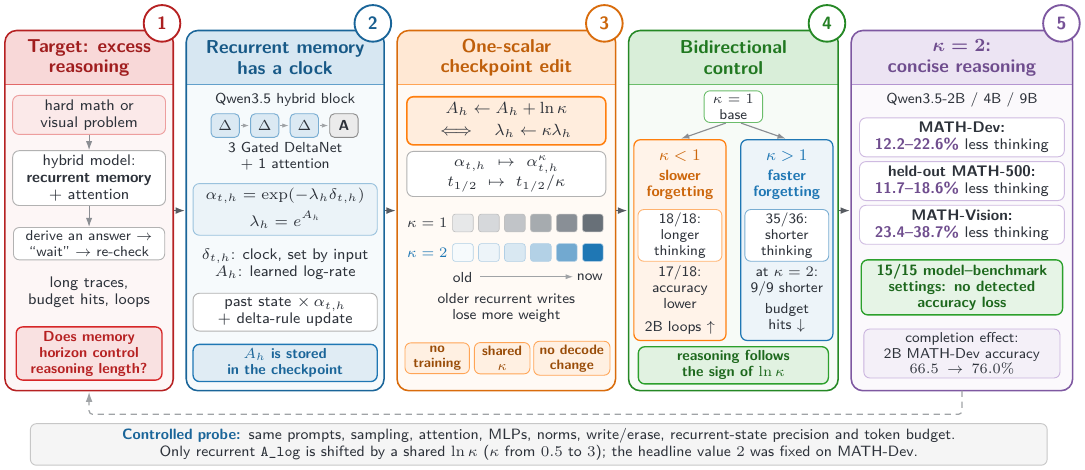}
  \caption{Overview of the study. (1)~Reasoning models often keep thinking
  after reaching an answer. (2)~In Qwen3.5, a learned log-rate $A_h$ per
  recurrent head, stored in the checkpoint, sets how fast memory decays.
  (3)~The edit adds $\ln\kappa$ to every $A_h$ and changes no other weight;
  half-lives are in clock time. (4)~Slower forgetting lengthened thinking and
  faster forgetting shortened it; counts are model--benchmark--$\kappa$
  settings for $\kappa$ from 0.5 to 2. (5)~At
  $\kappa=2$, fixed on MATH-Dev, thinking fell on every benchmark with no
  detected loss of accuracy.}
  \label{fig:overview}
\end{figure*}

Cheaper tokens do not help much if a model writes far more of them than it
needs. Reasoning models often keep going after they have found the answer,
re-checking results that are already settled \citep{chen2024overthinking}.
One analysis finds that 52--88\% of chain-of-thought tokens are produced
after the answer is already recoverable from a partial prefix
\citep{wang2026detection}, and long traces sometimes fall into repetitive
loops \citep{pipis2025loop}. Such over-reasoning raises cost without a
matching gain in accuracy, while reasoning that stops too early leaves
solutions incorrect or incomplete \citep{mia2026adaptive}. Existing methods retrain the model with
length-aware rewards \citep{arora2025efficient}, change decoding so that it
exits early or suppresses reflection tokens 
\citep{yang2025deer,wang2025nowait}, or push activations along directions
learned from data \citep{chen2025seal,azizi2025asc}; related weight edits
can also lengthen reasoning that is too short \citep{sun2025thinkedit}.
None of them directly manipulates the decay of the recurrent memory, the
part that makes a hybrid different. Work on recurrent decay, for its part, has focused on context extension
and length generalization, including making recurrent models forget
\citep{azizi2025mambaextend,lu2025modulation,chen2024stuffed}, and
studies of hybrids have asked which component retrieves facts and which
shapes the style of the output \citep{afendulev2026recalls}. To our
knowledge, no earlier work has asked whether the decay rate of a hybrid's
recurrent layers affects how long it reasons.

We study this question with the simplest intervention we could find. We
multiply the decay rate of every recurrent head by one factor $\kappa$.
Within a layer, this raises every forget gate to the power $\kappa$ and
shrinks the half-life of every gate share by $\kappa$ when time is
measured in the head's own clock; the attention weights, prompts and
decoding are left unchanged (\S\ref{sec:method}). The edit touches one small tensor per recurrent
layer, uses no training or learned direction and, once $\kappa$ is chosen,
no data. In
Qwen3.5, each recurrent head's read-out is normalized on its own, which approximately absorbs the
overall drop in volume that faster decay would otherwise cause. The edit
belongs to a family of decay edits used to extend the context of recurrent
models; as released, those edits raise each rate to a power or fit their
factors to data, whereas ours multiplies every rate by one shared factor
set by hand (\S\ref{sec:prior-edits}). Our contribution is what this
simplest form reveals about reasoning. Figure~\ref{fig:overview}
summarizes the study: changing one stored scalar per recurrent head moves
reasoning length in both directions, and at $\kappa=2$ thinking becomes
shorter without a detected loss of accuracy. We vary $\kappa$ from 0.5 to 3 in both directions on Qwen3.5
models from 2B to 9B parameters and measure accuracy, reasoning length,
budget overruns, loops and re-checking on MATH problems of levels 3--5
\citep{hendrycks2021math}, GSM8K \citep{cobbe2021gsm8k} and AIME 2024 and
2025. We report every value of $\kappa$, with main
results at one fixed value, $\kappa=2$. We then test this value on held-out MATH-500 problems
\citep{lightman2024verify} and on visual problems from MATH-Vision
\citep{wang2024mathvision}, where we also compare it with suppressing
reflection tokens \citep{wang2025nowait}.

\paragraph{Contributions.}
\begin{itemize}
  \setlength{\itemsep}{1pt}\setlength{\parsep}{0pt}\setlength{\topsep}{2pt}
  \item \textbf{Recurrent memory decay controls reasoning length.} In
    Qwen3.5 hybrids, reasoning length follows the decay rate of the
    recurrent memory in both directions: faster forgetting shortens it and
    slower forgetting lengthens it, in all but one of the 54
    model--benchmark--$\kappa$ settings from $\kappa=0.5$ to 2, with only
    the decay rates of the recurrent layers changed.

  \item \textbf{A minimal probe with stated assumptions.} We use the
    simplest decay edit, one shared shift of every head's log decay rate,
    as a probe. We collect the properties that hold exactly within a layer
    with its inputs fixed, such as each forget gate rising to the power
    $\kappa$, most of which follow from known facts about gated
    recurrences, and separate them from what must be measured in the full
    network. The edited checkpoint runs in a standard serving stack with no
    additional operations at inference.
  \item \textbf{Shorter reasoning with no detected loss of accuracy.}
    Across three Qwen3.5 models (2B, 4B and 9B) and three benchmarks
    (MATH-Dev, GSM8K and AIME), the edit at $\kappa=2$ shortened thinking
    in all nine settings, by 12.2--22.6\% on
    MATH-Dev, and no setting showed a detected drop in accuracy. On
    held-out MATH-500 it shortened thinking by 11.7--18.6\%, with no
    detected drop in accuracy, and on MATH-Vision by 23.4--38.7\%. Under a fixed token budget, accuracy rose
    from 66.5\% to 76.0\% for the 2B model on MATH-Dev, mainly because more
    of its traces finished in time. Slower forgetting had the opposite
    effect: thinking grew by up to 61\% and accuracy fell in 17 of 18
    settings.
\end{itemize}
\section{Related Work}
\label{sec:related}

\paragraph{Overthinking and efficient reasoning.} Reasoning models trained with reinforcement learning produce long chains of thought \citep{deepseek2025r1} and gain accuracy from more test-time computation \citep{snell2024scaling}, but much of it is redundant: models re-check results they have already reached \citep{chen2024overthinking}, a large share of the trace is written after the answer can already be recovered from it \citep{wang2026detection}, and long traces can fall into loops \citep{pipis2025loop}. Existing methods fall into three groups. Training-based methods reward short correct traces with reinforcement learning \citep{arora2025efficient}. Decoding-time methods stop generation once the model appears confident in a trial answer \citep{yang2025deer}, enforce a thinking budget \citep{muennighoff2025s1}, or suppress reflection tokens such as ``wait'' \citep{wang2025nowait}. Representation-level methods add or remove directions, in the residual stream or in selected attention heads, that are associated with reflection or verbosity \citep{chen2025seal,azizi2025asc,huang2025manifold,zhang2025crest}. These methods act on the model's rewards, output tokens or activations, and most need training, curated data or a controller tuned to each model; none of them directly manipulates recurrent memory decay.

\paragraph{Weight edits and reasoning length.} A smaller line of work changes reasoning length by editing weights. ThinkEdit removes a short-reasoning direction from the output projections of a few attention heads to lengthen overly short reasoning \citep{sun2025thinkedit}, and Steer2Edit turns steering vectors into component-level weight edits and reports that, at matched downstream performance, these edits outperform the strongest steering baseline by 12.2\% in reasoning-length reduction \citep{sun2026steer2edit}. Weight changes made for other purposes also move reasoning length: low-bit quantization makes reasoning models write longer chains of thought \citep{lian2026inflates,lotfi2026quantized}. These edits either follow a direction learned from data or arise as a side effect of compression. Our edit changes a single architectural quantity, the decay rate of the recurrent memory, by the same factor in every head; it needs no learned direction and, once the factor is chosen, no data.

\paragraph{Recurrent and hybrid language models.} State-space models such as Mamba \citep{gu2023mamba} and Mamba-2 \citep{dao2024mamba2}, and gated linear-attention models with delta-rule updates such as Gated DeltaNet \citep{yang2025gdn}, replace the growing key--value cache with a fixed-size recurrent state whose contents decay at a rate set by a learned decay parameter and an input-dependent step. Hybrid models combine such layers with attention, either interleaved across layers, as in Jamba \citep{lieber2024jamba}, Nemotron-H \citep{nvidia2025nemotronh} and Qwen3.5 \citep{qwen35}, or in parallel within each layer, as in Falcon-H1 \citep{falcon2025h1}. Several hybrids are now trained as reasoning models \citep{nvidia2025nemotronnano2,falcon2026h1r}. Analyses of hybrids find that exact retrieval from the context runs through attention while the recurrent state shapes properties such as output language and persona \citep{afendulev2026recalls}, and compare hybrid and non-hybrid models on state tracking and recall \citep{rawat2026primitives}. For decoders of fixed depth and constant precision, \citet[Theorem~4.2]{steifer2026scratchpads} proves a separation on a parity-conditioned retrieval task: a three-layer hybrid of Gated DeltaNet and gated attention with logarithmic width solves it without a scratchpad, pure gated attention of logarithmic width needs a polynomial-length scratchpad, and pure Gated DeltaNet fails with any scratchpad unless its rounded state holds at least $n-1$ bits for $n$ input bits. Concurrent work finds that storing the recurrent state in bfloat16 makes hybrid reasoning models, including Qwen3.5-9B, reason longer \citep{weng2026rounding}; it changes the numerical precision of the state, whereas we keep the state in float32 and change the decay rate itself.

\paragraph{Controlling memory decay.} How fast recurrent memory forgets has so far been studied as a long-context problem. \citet{chen2024stuffed} trace the failure of Mamba models on inputs longer than their training length to an inability to forget: trained on contexts too short for their state size, the models never need to learn to forget. MambaExtend repairs such failures by calibrating a per-layer scale on the discretization step; since the step enters both the decay and the input term, this scale changes both \citep{azizi2025mambaextend}. Mamba Modulation instead scales $A$; its released code applies the factor to $\log A$, which raises each decay rate to a power \citep{lu2025modulation}. As a first test, it applies one untuned constant, 2 or 3, in every layer, which kept the long-context perplexity of a Mamba-2 model from exploding, although the authors judge constant scaling generally ineffective; its main method calibrates one factor per layer. Other training-free methods act on what enters the state rather than on its decay: LongMamba accumulates only selected tokens in the channels with long memory \citep{ye2025longmamba}, and DeciMamba removes tokens with small steps in selected layers \citep{benkish2025decimamba}. All of these target pure Mamba models on inputs longer than those seen in training. We instead multiply every decay rate by one factor, shared by all heads and set by hand, which divides every gate half-life by the same amount (\S\ref{sec:prior-edits} compares the forms), and apply it to Qwen3.5 hybrid reasoning models in both directions, on traces well inside their training context, to ask how memory decay shapes the length of reasoning.
\section{Methodology}
\label{sec:method}

We define a family of checkpoint edits $E_\kappa$, $\kappa>0$, each of
which changes how fast every recurrent head forgets and no other weight. An
edit is applied once, before serving, and adds no operations at inference.
$E_1$ leaves the model as it is, so the effect can be studied as a function
of $\kappa$ alone.

\paragraph{Forget gates in Gated DeltaNet.}
Each head $h$ of a Gated DeltaNet layer carries a matrix state, updated as
\begin{equation}
  S_t = \alpha_{t,h}\,S_{t-1}\big(I-\beta_t k_t k_t^{\top}\big)
        + \beta_t v_t k_t^{\top},
  \label{eq:gdn}
\end{equation}
where $\beta_t$ is the write strength and $\alpha_{t,h}\in(0,1)$ is the
forget gate; we drop the head index on $\beta$, $k$ and $v$. The update
combines the delta rule of fast-weight programmers and DeltaNet
\citep{schlag2021fwp,yang2024deltanet} with the decay gate of gated linear
attention and Mamba-2 \citep{yang2024gla,dao2024mamba2}, as introduced by
Gated DeltaNet \citep{yang2025gdn}. Qwen-style models compute the gate as
\begin{equation}
  \alpha_{t,h} = \exp(-\lambda_h\,\delta_{t,h}),
  \qquad \lambda_h = e^{A_h},
  \label{eq:gate}
\end{equation}
where $A_h$ is a learned scalar per head, stored as \texttt{A\_log}, and
$\delta_{t,h}=\mathrm{softplus}(a_{t,h}+b_h)>0$ is a step that depends on
the input through a projection $a_{t,h}$ and a bias $b_h$. Mamba-2 uses the
same gate with its own step $\Delta_{t,h}$ in place of $\delta_{t,h}$
\citep{dao2024mamba2}. It helps to think of $\delta$ as the head's clock
and of $\lambda_h$ as how much it forgets per tick. Reading a gate as a
learned time constant, and a change of its rate as a rescaling of time,
goes back to work on gated recurrent networks
\citep{tallec2018warp,mahto2021timescales}, and linear recurrent units
parameterize their decay in the same doubly exponential form
\citep{orvieto2023lru}. If
$T_h(s,t)=\sum_{u=s+1}^{t}\delta_{u,h}$ is the clock time between tokens
$s$ and $t$, the gate keeps a share
\begin{equation}
  R_h(s,t) = \prod_{u=s+1}^{t}\alpha_{u,h} = e^{-\lambda_h T_h(s,t)}
  \label{eq:retention}
\end{equation}
of a value written at $s$. The erase term in Eq.~\ref{eq:gdn} can only
remove more: in Qwen3.5 the write strength is a sigmoid and keys are
$L_2$-normalized, so $I-\beta_t k_t k_t^{\top}$ never increases the norm,
and $R_h$ is an upper bound on how much of the value survives, not an
exact measure. The rate $\lambda_h$ and the bias $b_h$ are both fixed after
training; only the projection $a_{t,h}$ depends on the input. How many
tokens a head keeps a value therefore depends on its rate and its clock
together, so two heads with equal $\lambda_h$ can keep values for very
different numbers of tokens. The edit below changes only $\lambda_h$.

\paragraph{The edit.}
For every recurrent head in every recurrent layer we set
\begin{equation}
  A_h \leftarrow A_h + \ln\kappa
  \quad\Longleftrightarrow\quad
  \lambda_h \leftarrow \kappa\,\lambda_h .
  \label{eq:edit}
\end{equation}
With $\kappa>1$ the model forgets faster, and with $\kappa<1$ more slowly.
Only one scalar per recurrent head changes: 288 of them in Qwen3.5-2B
(18 Gated DeltaNet layers of 16 heads) and 768 in Qwen3.5-4B and 9B (24
layers of 32 heads). Every other weight stays as it was. Since the edit
adds to $A_h$, its natural scale is $\ln\kappa$, and $\kappa$ and
$1/\kappa$ are equal steps in opposite directions.
\begin{figure}[t]
  \centering
  \includegraphics[width=0.85\columnwidth]{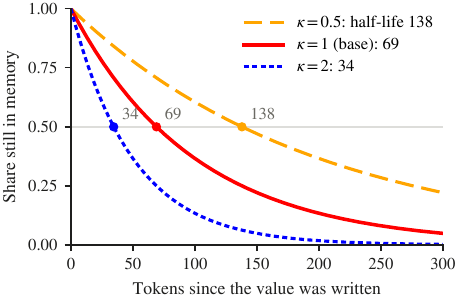}
  \caption{Gate share $R_h$ of a value written $n$ tokens earlier, for an
  illustrative constant gate $\alpha=0.99$, not a value measured in a
  model. The edit divides the half-life of this share by $\kappa$; dots
  mark the half-lives.}
  \label{fig:halflife}
\end{figure}

\paragraph{What holds exactly within a layer.}
For fixed gate inputs, the edit has four exact properties:
\begin{itemize}
  \setlength{\itemsep}{0pt}\setlength{\parsep}{0pt}\setlength{\topsep}{2pt}
  \item[(P1)] \emph{Power law.} Each gate becomes $\alpha_{t,h}^{\kappa}$,
    so each share $R_h$ becomes $R_h^{\kappa}$.
  \item[(P2)] \emph{Half-life of the gate share.} Measured in clock time
    $T_h$, the half-life of every share $R_h$ shrinks by exactly $\kappa$.
    Measured in tokens, it shrinks by exactly $\kappa$ only when the step
    $\delta$ is constant, where the half-life becomes
    $\ln 2/(\kappa\lambda_h\bar\delta)$ (Figure~\ref{fig:halflife};
    Proposition~\ref{prop:power}); otherwise it cannot grow but can
    shrink by much more or less than $\kappa$. The erase term,
    which the edit leaves unchanged, removes a further part of each value,
    so for $\kappa\ge 1$ the half-life of what actually survives shrinks by
    a factor between $1$ and about $\kappa$: close to $\kappa$ when
    erasure is weak, and less when it dominates.
  \item[(P3)] \emph{Ratios kept.} All log-rates move by the same
    $\ln\kappa$, so the ratios $\lambda_h/\lambda_{h'}$ do not change, and
    neither does the order of any two heads' gate shares $R_h(s,t)$ and
    $R_{h'}(s,t)$ between the same two tokens.
  \item[(P4)] \emph{Gate validity and contraction.} Gates stay in $(0,1)$ for every
    $\kappa>0$, and edits compose:
    \mbox{$E_\kappa\circ E_{\kappa'}=E_{\kappa\kappa'}$}. With
    $\kappa\ge 1$ the gates only shrink, so before the per-head norm the
    contribution of each past write can only fall, although the norm of the
    whole state need not, since writes of opposite sign can cancel less.
    After the norm, the edit re-weights rather than shrinks (see below).
\end{itemize}
These properties describe one layer with its inputs held fixed (proofs in
Appendix~\ref{app:proofs}). In the full network, only the first recurrent
layer receives the same inputs in both runs, and only when the same text is
fed in; every later layer sees hidden states changed by the edit, and during
generation the text itself differs, so the step $\delta$, the write
strength, the keys and the values all differ from the unedited run. The
effect on reasoning therefore has to be measured, and
\S\ref{sec:experiments} does so.

\paragraph{What stays the same.}
The weights that produce the clock $\delta_{t,h}$, which decides
\emph{when} a head forgets, are untouched, as are those for the write
strength, the delta-rule erase, the attention layers, MLPs and norms, and
the prompts and decoding. The edit changes only the recurrent decay-rate
parameters. Functionally, it changes the time scale of the recurrent layers
and re-weights older against newer contributions to their state. The
attention layers, one in four in Qwen3.5, keep full-context connectivity,
although the representations they receive can change downstream of the
edit. Because $\beta_t$ does not depend on
$\delta_{t,h}$, scaling the step by $\kappa$ and shifting $A_h$ give
exactly the same model in Gated DeltaNet. Mamba-2 is different, since there
the step also multiplies the write (see below).

\paragraph{Effect on the read-out scale.}
Faster decay also makes a head quieter. In a simplified picture with a
constant gate, ignoring the erase term and the content of the writes, the
read-out weights a value written $n$ tokens ago by $\alpha^n$, for a
geometric gain of $G(\alpha)=1/(1-\alpha)$; after the edit
$G(\alpha^\kappa)\approx G(\alpha)/\kappa$ for long-memory heads
($\alpha\to 1$), while short-memory heads ($\alpha\to 0$) stay near
$G\approx 1$. Qwen-style models pass each head's read-out through its own
RMS norm \citep{zhang2019rmsnorm} before the output gate and projection,
with no skip term (as implemented in Transformers 5.20.2,
\citealp{wolf2020transformers}), so this drop in volume is normalized away,
up to the small constant $\epsilon$ inside the norm. The edit still changes
the direction of the read-out, because it re-weights past tokens: the newest
write's share of the total weight grows from $1-\alpha$ to
$1-\alpha^{\kappa}$, by a factor that approaches $\kappa$ for long-memory
heads (Appendix~\ref{app:proofs}). This re-weighting is the intended effect,
and we do not claim that the edit changes nothing but the time scale of the
recurrent layers. Qwen3-Next uses the same gate and per-head norm, so the
edit carries over to it unchanged; in Mamba-2-based hybrids that share one
norm across heads or add a skip term, it would also shift the balance
between heads.

\paragraph{Relation to earlier decay edits.}
\label{sec:prior-edits}
Two earlier works rescale decay without retraining the model, both to
extend the context of pure Mamba models. MambaExtend 
\citep{azizi2025mambaextend} calibrates one factor per layer on the step
$\Delta$, which in Mamba also scales the write. Mamba Modulation
\citep{lu2025modulation} describes its factors as a scaling of $A$
without stating the update. Its released code applies
the factor to the log-rate, $A_h\leftarrow sA_h$ in our notation, so each
rate becomes $\lambda_h^{s}$ rather than $\kappa\lambda_h$. This power law
is a different edit: with $s>1$ it makes heads with $\lambda_h>1$ forget
faster but heads with $\lambda_h<1$ forget more slowly, and it changes the
ratios that (P3) keeps. The paper uses it in two ways: as a first test, one
untuned constant, 2 or 3, shared by all layers, and, as its main method,
one factor per layer calibrated on data. Only its Mamba-2 passkey
experiments multiply each head's rate, as Eq.~\ref{eq:edit} does, but by a
separate factor per head, trained on retrieval data while all other weights
stay frozen.%
\footnote{\raggedright Commit \texttt{0ac3cb3} of
\url{https://github.com/gnepul-ace/mamba_modulation}. In
\nolinkurl{Mamba2/perplexity_eval/modeling/mamba_module.py}, line~185,
the rate is $e^{sA_h}$, as in the Mamba-2 LongBench module and all Mamba-1
modules; in \nolinkurl{Mamba2/PasskeyRetrieval/modeling/mamba_module.py},
line~226, it is $s_h e^{A_h}$. We found no separate script for the
constant test.}

As implemented in released code, then, no earlier edit multiplies every
decay rate by one shared factor that is not fitted to data. Because the
text of Mamba Modulation does not state its update, its constant test could
also be read as Eq.~\ref{eq:edit} with $\kappa=2$ or 3; even under that
reading, it was applied to pure Mamba models to extend their context.
$E_\kappa$ differs from these uses in four ways. (i)~\emph{Form}: one
factor multiplies every rate in every head and layer, rather than a power
of each rate or a factor fitted per layer or per head.
(ii)~\emph{Architecture}: the edit acts on Gated DeltaNet layers in a hybrid
backed by attention, where each head is normalized on its own and where
scaling the rate and scaling the step coincide, rather than on a pure Mamba
model. (iii)~\emph{Purpose}: we study reasoning length on inputs well
inside the training context, rather than retrieval or perplexity beyond it.
(iv)~\emph{Range}: we vary $\kappa$ in both directions, from 0.5 to 3, and
report every value we ran; only the main value, $\kappa=2$, was chosen on a
development set (\S\ref{sec:setup}). Only a shared factor on the rate
keeps all ratios between heads, within and across layers (P3), and gives a
one-parameter family that composes and inverts (P4). The family of decay
edits is not new; our contribution is this shared, unfitted form as a
probe of reasoning length, and what it reveals.
\paragraph{Implementation.}
Each $\kappa$ setting uses an overlay checkpoint that links to the original weight
shards and replaces only the \texttt{A\_log} tensors, shifted in float32
without loading the model, and vLLM \citep{kwon2023efficient} serves it
unchanged. Before generating, we check that the edit finds the expected
layers and heads and that the engine reads the edited tensors: in the
evaluation configuration, the $\kappa=1$ overlay reproduces the base
log-probabilities exactly, while the $\kappa=2$ overlay changes them by
0.15--0.19 nats per token (Appendix~\ref{app:repro}).
\section{Experiments}
\label{sec:experiments}

\begin{figure*}[t]
  \centering
  \includegraphics[width=0.85\textwidth]{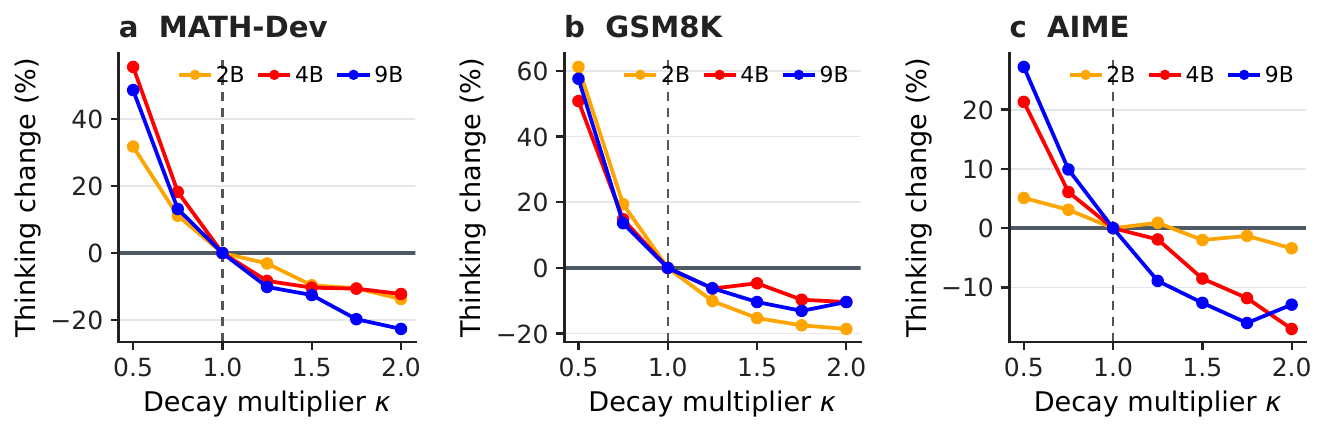}
  \caption{Change in mean thinking tokens from the unedited model (\%) for
    each value of $\kappa$ from 0.5 to 2, for each benchmark and model. The dashed vertical
    line marks the unedited model ($\kappa=1$); $\kappa<1$ forgets more
    slowly and $\kappa>1$ faster. Exact values are in Table~\ref{tab:main}
    (Appendix~\ref{app:full}).}
  \label{fig:tokens}
\end{figure*}

\subsection{Setup}
\label{sec:setup}

\paragraph{Models and data.}
We edit Qwen3.5-2B, Qwen3.5-4B and Qwen3.5-9B \citep{qwen35}.
\emph{MATH-Dev} has 200 problems of levels 3--5 from
the test split of MATH \citep{hendrycks2021math}; we remove the 500 problems
of MATH-500 \citep{lightman2024verify}, which we hold out as a test set
(\S\ref{sec:heldout}), and every near-duplicate of a MATH-500 or AIME
problem before sampling. We also use the 1,319 test problems of GSM8K
\citep{cobbe2021gsm8k}, the 60 problems of AIME 2024 and 2025 and, in
\S\ref{sec:vision} only, the 304 problems of MATH-Vision testmini
\citep{wang2024mathvision}. Appendix~\ref{app:repro} gives all further
details.

\paragraph{Compared experimental settings.}
We compare the unedited model ($\kappa=1$; the base model in tables) with
two edits that forget more slowly ($\kappa=0.5$, 0.75) and four that forget
faster ($\kappa=1.25$, 1.5, 1.75, 2). For headline numbers we use one value for every model and benchmark,
$\kappa=2$, which we fixed on MATH-Dev before running the held-out experiments,
rather than the most accurate value in each setting. We chose it by hand: on
MATH-Dev, every value from 1.25 to 2 shortened thinking for every model and
raised or kept accuracy (Table~\ref{tab:main}). Results for $\kappa=2.25$ to 3
are in Appendix~\ref{app:above2}.

\paragraph{Decoding and measures.}
All settings use the model card's thinking-mode sampling parameters under a budget
of 34,000 generated tokens, with one response per question and a seed
derived from the question, so every setting starts from the same random state.
A response that does not close its thinking within the budget counts as
wrong. We report accuracy, mean thinking tokens (tokens before the
end-of-thinking tag), the share of responses that reach the budget, and
accuracy among finished responses. Token changes are relative to the
unedited model and accuracy changes are in percentage points (pp).
Intervals are 95\% paired bootstrap intervals over questions, not corrected
for multiple comparisons; an interval that includes zero means no change was
detected, not that the two settings are equivalent.

\subsection{Main results}
\label{sec:main}

\begin{figure*}[t]
  \centering
  \includegraphics[width=0.85\textwidth]{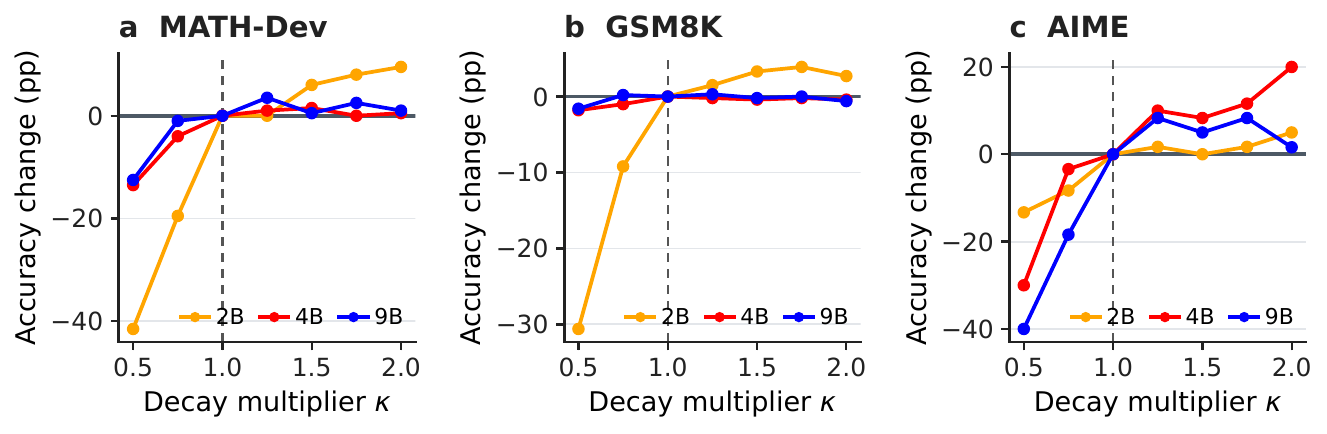}
  \caption{Change in accuracy from the unedited model (pp) for each value of
    $\kappa$ from 0.5 to 2, for each benchmark and model. The dashed vertical line marks
    the unedited model ($\kappa=1$). Responses that reach the 34,000-token
    budget count as wrong. Exact values are in Table~\ref{tab:main}
    (Appendix~\ref{app:full}).}
  \label{fig:acc}
\end{figure*}
\begin{table*}[t]
  \centering\small
  \begin{tabular}{@{}llcccc@{}}
    \toprule
    Model & Benchmark & $\Delta$Thinking (\%) & $\Delta$Acc. (pp) & Budget hits (\%) & Fin. acc. (\%) \\
          & & [95\% CI] & [95\% CI] & base$\to\kappa{=}2$ & base$\to\kappa{=}2$ \\
    \midrule
    2B & MATH-Dev & \cellcolor{bestcell}$-13.7$ [$-18.3$, $-9.0$] & \cellcolor{bestcell}$+9.5$ [3.5, 15.5] & 32.0$\to$20.0 & 97.8$\to$95.0 \\
       & GSM8K & \cellcolor{bestcell}$-18.6$ [$-21.3$, $-15.7$] & \cellcolor{bestcell}$+2.7$ [0.8, 4.5] & 13.4$\to$4.9 & 95.5$\to$89.7 \\
       & AIME & $-3.4$ [$-7.7$, 0.4] & $+5.0$ [$-1.7$, 13.3] & 86.7$\to$75.0 & 100.0$\to$73.3 \\
    \midrule
    4B & MATH-Dev & \cellcolor{bestcell}$-12.2$ [$-17.7$, $-6.2$] & $+0.5$ [$-1.5$, 3.0] & 3.0$\to$2.0 & 97.9$\to$97.4 \\
       & GSM8K & \cellcolor{bestcell}$-10.4$ [$-15.4$, $-4.9$] & $-0.4$ [$-1.4$, 0.6] & 0.8$\to$0.4 & 96.5$\to$95.7 \\
       & AIME & \cellcolor{bestcell}$-17.0$ [$-22.1$, $-12.1$] & \cellcolor{bestcell}$+20.0$ [10.0, 31.7] & 46.7$\to$21.7 & 96.9$\to$91.5 \\
    \midrule
    9B & MATH-Dev & \cellcolor{bestcell}$-22.6$ [$-27.6$, $-17.2$] & $+1.0$ [$-1.5$, 4.0] & 3.0$\to$1.5 & 99.0$\to$98.5 \\
       & GSM8K & \cellcolor{bestcell}$-10.4$ [$-15.1$, $-5.3$] & $-0.6$ [$-1.4$, 0.2] & 0.1$\to$0.3 & 96.4$\to$96.0 \\
       & AIME & \cellcolor{bestcell}$-12.9$ [$-18.7$, $-7.0$] & $+1.7$ [$-8.3$, 13.3] & 31.7$\to$30.0 & 97.6$\to$97.6 \\
    \bottomrule
  \end{tabular}
  \caption{The edit at $\kappa=2$ against the unedited model. Changes are paired by question, with 95\% bootstrap intervals. 2B/4B/9B: Qwen3.5-2B/4B/9B. Fin.\ acc.: accuracy among finished responses, descriptive since different questions finish in each setting. Shaded: the change favours the edit (less thinking or higher accuracy) and its 95\% interval excludes zero.}
  \label{tab:k2}
\end{table*}
\paragraph{Faster forgetting shortens reasoning.}
Edits with $\kappa>1$ shortened mean thinking in 35 of the 36
model--benchmark--$\kappa$ settings, and in 29 of them the interval lies
below zero (Figure~\ref{fig:tokens}). The exception is the 2B model on AIME
at $\kappa=1.25$ ($+0.9\%$ [$-2.5$, $4.2$]); without the edit this model
reaches the budget on 86.7\% of AIME problems. At $\kappa=2$
(Table~\ref{tab:k2}), thinking fell in all nine settings, and every interval lies below zero except for the 2B model on
AIME. Thinking plus answer tokens fell by similar amounts
(3.1--21.7\%), so the answer does not grow to compensate.

At $\kappa=2$, accuracy (Figure~\ref{fig:acc}) rose
in seven of the nine settings and fell slightly in the other two, by
0.4~pp for the 4B model and 0.6~pp for the 9B model, both on GSM8K; neither
interval lies below zero. No setting shows a detected loss of accuracy at
any value of $\kappa$ from 1.25 to 2. The intervals lie above zero for the
2B model on MATH-Dev (66.5\% to 76.0\%) and GSM8K ($+2.7$~pp) and for the 4B
model on AIME ($+20.0$~pp, 12 of 60 problems).

\paragraph{Slower forgetting lengthens it.}
The effect reverses with the sign of $\ln\kappa$ (Figures~\ref{fig:tokens}
and~\ref{fig:acc}). Every edit with $\kappa<1$
lengthened thinking in every setting: at $\kappa=0.5$, by 31.7\%, 55.4\%
and 48.5\% on MATH-Dev and by 50.8--61.2\% on GSM8K. Accuracy fell in 17 of
these 18 settings, for example from 66.5\% to 25.0\% for the 2B model on
MATH-Dev; the exception is the 9B model on GSM8K at $\kappa=0.75$
($+0.2$~pp, interval includes zero). The 2B model's traces degenerated:
looping traces rose from 13 to 87 on MATH-Dev and from
47 to 455 on GSM8K. The size of the change in thinking, however, did not
always grow with $\kappa$, most clearly where traces reach
the budget at every value (Appendix~\ref{app:monotone}).
Reasoning length thus follows the decay rate in both directions. That the
effect flips with the sign of $\ln\kappa$ argues against it being a
by-product of perturbing the checkpoint at all, although an edit with random
signs would test this directly. At $\kappa=2$, by
contrast, looping traces rose or fell only slightly and the density of
re-check phrases barely changed (Appendix~\ref{app:loops}).

\subsection{Held-out test on MATH-500}
\label{sec:heldout}

To test whether the results carry over to unseen problems, we ran the base
model and the edit at $\kappa=2$ on MATH-500, with all other settings as in
\S\ref{sec:setup}. MATH-500 played no part in
choosing $\kappa$ or any other setting and covers difficulty levels
1--5.

\paragraph{The effect holds on held-out problems.}
The edit shortened thinking for all three models, by 11.7\%, 17.2\% and
18.6\%, with every interval below zero (Table~\ref{tab:heldout}). Accuracy rose for all
three models; the interval lies above zero for the 2B model ($+10.0$~pp)
and includes zero for the 4B and 9B models (both $+1.6$~pp). The saving
depends on difficulty: every interval lies below zero on levels 3--5, the
levels of MATH-Dev, and the 9B model saved 6.4\% on level~1 but 21.6\% on
level~5 (Appendix~\ref{app:heldout}). At 81,920 tokens the MATH-500
saving was 19.6--24.1\%, again with no detected change in accuracy
(Appendix~\ref{app:budget82k}).

\paragraph{Where the accuracy gain comes from.}
Since unfinished responses count as wrong, accuracy mixes completion with
correctness. Following each MATH-500 question from the base model to the
edit (Table~\ref{tab:heldout}), most of the 2B model's net gain of 50
questions comes from responses that now finish: the edit finished 79 of the
161 questions the base model left unfinished, and the model's accuracy among
finished responses fell from 97.6\% to 94.1\% (Appendix~\ref{app:heldout}).
We therefore read the accuracy gains as a benefit under this budget and the
shorter thinking as the main result.

\subsection{Visual mathematical reasoning}
\label{sec:vision}

Qwen3.5 models take images as input, so we also ran the base model, the
edit at $\kappa=2$ and NoWait on MATH-Vision; NoWait
\citep{wang2025nowait} suppresses reflection tokens such as ``wait''.

The edit shortened thinking by 38.7\%, 23.4\% and 24.7\% for the 2B, 4B and
9B models, more than on any text benchmark, with every interval below zero
(Table~\ref{tab:vision}). Few base responses reach the budget here (8.6\%,
0.3\% and 0.3\%), so the saving does not come from cutting overruns.
Accuracy changed by $-3.9$, $+0.3$ and $-0.7$~pp, every interval including
zero, although the 2B model's interval ($[-9.9, 2.0]$) leaves room for a loss of
several points and its loss on level~2 was detected (Appendix~\ref{app:vision}). On
the same questions, the edit thought 26.0\% less than NoWait for the 2B
model and 7.4\% less for the 9B model, with no detected difference for the
4B model or in accuracy.

\section{Conclusion}
\label{sec:conclusion}

Recurrent memory decay controls how long the evaluated Qwen3.5 hybrids
reason, in both directions. With one shared shift of every head's log decay
rate, applied once with no training or decoding change, faster forgetting at
$\kappa=2$ shortened thinking in every setting, by 12.2--22.6\% on MATH-Dev,
by 11.7--18.6\% on held-out MATH-500 and by 23.4--38.7\% on MATH-Vision, with
no detected loss of accuracy in any of the fifteen model--benchmark settings
at our 34,000-token budget; slower forgetting lengthened it, almost always
lowered accuracy and, in the 2B model, made traces loop. The 2B model's
accuracy gains come mainly from traces that now finish in time.
\section*{Limitations}

Our runs cover Qwen3.5 models up to 9B parameters. Larger Qwen hybrids such
as Qwen3-Next use the same gate and the same per-head norm, so the edit
applies to them as is, but we have not measured them. Hybrids built on
Mamba-2, or with attention running beside the recurrent layer in every
block, normalize their heads differently (\S\ref{sec:method}), and the
effect there may differ. All tasks are mathematical, and we did not test
whether the edit changes performance on other tasks.

We show that changing the decay changes reasoning length, but not why. The
density of re-check phrases barely changed (Appendix~\ref{app:loops}), and this
lexical proxy does not show which reasoning steps the edit removes. Edits of
the same size with random signs, edits restricted to some layers and a
side-by-side reading of paired traces would test which part of the change
matters.

We fixed $\kappa=2$ on MATH-Dev, a 200-problem development set, so the
MATH-Dev results for $\kappa=2$ are in-sample, and tested
it on one held-out set, MATH-500, from the same benchmark; GSM8K and AIME
have no held-out counterpart, and all values of $\kappa$ were examined on
them. The useful range is limited: above $\kappa=2$ the mean saving grew
little while mean accuracy fell, with detected losses at $\kappa=2.25$ and 3
(Appendix~\ref{app:above2}). We drew one response per question in a single run, so
the intervals capture variation across questions but not across samples or
runs; the 60 AIME problems give wide intervals, and the intervals are not
corrected for multiple comparisons.

Our budget of 34,000 tokens is shorter than the one the model card
recommends for competition mathematics, and traces that exceed it count as
wrong. Most of the accuracy gain comes from responses that now finish,
while accuracy among finished responses fell in eight of the nine
development settings (Table~\ref{tab:k2}); with a budget of 81,920 tokens the
saving on MATH-Dev and MATH-500 remained, with no detected change in
accuracy (Appendix~\ref{app:budget82k}). On MATH-Vision we ran only $\kappa=2$ on 304
problems; the 2B model's interval leaves room for a loss of several points,
its loss on level~2 was detected, and we do not know why the saving is
larger on visual problems.

We do not compare the edit with methods that need training or a direction
learned from data \citep{arora2025efficient,azizi2025asc}, since our
question is whether memory decay controls reasoning length; we make no
claim about how the edit ranks against them, and whether they can be
combined is untested. We count only generated tokens and do not measure
wall-clock time or memory.
\section*{LLM Usage Statement}

We used large language models (LLMs) to assist with manuscript preparation and code development. We used Claude (Anthropic) to edit manuscript drafts, analyze the results file generated by our experiments, and update our plotting script with the final numerical values. Claude also assisted in drafting, revising, and checking the proofs in the supplementary material. Before submission, we obtained reviewer-style feedback on the manuscript from ChatGPT (OpenAI) and Claude. We designed and conducted all experiments, verified all reported results, proofs, and citations, and take full responsibility for the content of this paper.

\bibliography{custom}

@article{deepseek2025r1,
  title={{DeepSeek-R1}: Incentivizing reasoning capability in {LLMs} via reinforcement learning},
  author={Guo, Daya and Yang, Dejian and Zhang, Haowei and Song, Junxiao and Wang, Peiyi and Zhu, Qihao and Xu, Runxin and Zhang, Ruoyu and Ma, Shirong and Bi, Xiao and others},
  journal={arXiv preprint arXiv:2501.12948},
  year={2025}
}

@article{snell2024scaling,
    title={Scaling {LLM} test-time compute optimally can be more effective than scaling model parameters},
  author={Snell, Charlie and Lee, Jaehoon and Xu, Kelvin and Kumar, Aviral},
  journal={arXiv preprint arXiv:2408.03314},
  year={2024}
}

@inproceedings{muennighoff2025s1,
    title = "s1: Simple test-time scaling",
    author = "Muennighoff, Niklas  and
      Yang, Zitong  and
      Shi, Weijia  and
      Li, Xiang Lisa  and
      Fei-Fei, Li  and
      Hajishirzi, Hannaneh  and
      Zettlemoyer, Luke  and
      Liang, Percy  and
      Cand{\`e}s, Emmanuel  and
      Hashimoto, Tatsunori",
    editor = "Christodoulopoulos, Christos  and
      Chakraborty, Tanmoy  and
      Rose, Carolyn  and
      Peng, Violet",
    booktitle = "Proceedings of the 2025 Conference on Empirical Methods in Natural Language Processing",
    month = nov,
    year = "2025",
    address = "Suzhou, China",
    publisher = "Association for Computational Linguistics",
    url = "https://aclanthology.org/2025.emnlp-main.1025/",
    doi = "10.18653/v1/2025.emnlp-main.1025",
    pages = "20275--20321",
    ISBN = "979-8-89176-332-6"
}

@article{arora2025efficient,
  title   = {Training Language Models to Reason Efficiently},
  author  = {Arora, Daman and Zanette, Andrea},
  journal = {arXiv preprint arXiv:2502.04463},
  year    = {2025}
}

@inproceedings{yang2025deer,
 author = {Yang, Chenxu and Si, Qingyi and Duan, Yongjie and Zhu, Zheliang and Zhu, Chenyu and Li, Qiaowei and Chen, Minghui and Lin, Zheng and Wang, Weipinng},
 booktitle = {International Conference on Learning Representations},
 editor = {C. Vondrick and B. Hariharan and C. Raffel and L. Pinto and D. Yang and A. Faust},
 pages = {88170--88210},
 title = {Dynamic Early Exit in Reasoning Models},
 url = {https://proceedings.iclr.cc/paper_files/paper/2026/file/8df90a1440ce782d1f5607b7a38f2531-Paper-Conference.pdf},
 volume = {2026},
 year = {2026}
}

@article{wang2026detection,
  title={The Detection-Extraction Gap: Models Know the Answer Before They Can Say It},
  author={Wang, Hanyang and Zhu, Mingxuan},
  journal={arXiv preprint arXiv:2604.06613},
  year={2026}
}

@article{gu2023mamba,
  title={Mamba: Linear-time sequence modeling with selective state spaces},
  author={Gu, Albert and Dao, Tri},
  journal={arXiv preprint arXiv:2312.00752},
  year={2023}
}

@article{lieber2024jamba,
  title={Jamba: A hybrid {Transformer-Mamba} language model},
  author={Lieber, Opher and Lenz, Barak and Bata, Hofit and Cohen, Gal and Osin, Jhonathan and Dalmedigos, Itay and Safahi, Erez and Meirom, Shaked and Belinkov, Yonatan and Shalev-Shwartz, Shai and others},
  journal={arXiv preprint arXiv:2403.19887},
  year={2024}
}

@article{nvidia2025nemotronh,
title={{Nemotron-H}: A family of accurate and efficient hybrid {Mamba-Transformer} models},
  author={Blakeman, Aaron and Basant, Aarti and Khattar, Abhinav and Renduchintala, Adithya and Bercovich, Akhiad and Ficek, Aleksander and Bjorlin, Alexis and Taghibakhshi, Ali and Deshmukh, Amala Sanjay and Mahabaleshwarkar, Ameya Sunil and others},
  journal={arXiv preprint arXiv:2504.03624},
  year={2025}
}

@inproceedings{huang2025manifold,
 author = {Huang, Yao and Chen, Huanran and Ruan, Shouwei and Zhang, Yichi and Wei, Xingxing and Dong, Yinpeng},
 booktitle = {Advances in Neural Information Processing Systems},
 doi = {10.52202/085713-3430},
 editor = {D. Belgrave and C. Zhang and H. Lin and R. Pascanu and P. Koniusz and M. Ghassemi and N. Chen},
 pages = {102543--102568},
 publisher = {Curran Associates, Inc.},
 title = {Mitigating Overthinking in Large Reasoning Models via Manifold Steering},
 url = {https://proceedings.neurips.cc/paper_files/paper/2025/file/946827aadb9f7dbdb27f4a0c77dd1859-Paper-Conference.pdf},
 volume = {38, Main Conference},
 year = {2025}
}

@article{zhang2025crest,
  title={Understanding and steering the cognitive behaviors of reasoning models at test-time},
  author={Zhang, Zhenyu and Wu, Xiaoxia and Zhou, Zhongzhu and Wu, Qingyang and Zhang, Yineng and Ponnusamy, Pragaash and Subbaraj, Harikaran and Wang, Jue and Song, Shuaiwen Leon and Athiwaratkun, Ben},
  journal={arXiv preprint arXiv:2512.24574},
  year={2025}
}

@article{sun2026steer2edit,
title={{Steer2Edit}: From Activation Steering to Component-Level Editing},
  author={Sun, Chung-En and Yan, Ge and Wang, Zimo and Weng, Tsui-Wei},
  journal={arXiv preprint arXiv:2602.09870},
  year={2026}
}

@article{lian2026inflates,
  title={Quantization Inflates Reasoning: Token Inflation as a Hidden Cost of Low-Bit Reasoning Models},
  author={Lian, Xinyu and Krichene, Walid and Huang, Beichen and Tanaka, Masahiro and Ruwase, Olatunji and Zhang, Li and Zhang, Minjia},
  journal={arXiv preprint arXiv:2606.25519},
  year={2026}
}

@article{lotfi2026quantized,
  title={Quantized reasoning models think they need to think longer, but they do not},
  author={Lotfi, Sanae and Kirichenko, Polina and Li, Steven and Liu, Zechun},
  journal={arXiv preprint arXiv:2606.00206},
  year={2026}
}

@inproceedings{yang2025gdn,
 author = {Yang, Songlin and Kautz, Jan and Hatamizadeh, Ali},
 booktitle = {International Conference on Learning Representations},
 editor = {Y. Yue and A. Garg and N. Peng and F. Sha and R. Yu},
 pages = {29687--29707},
title = {Gated Delta Networks: Improving {Mamba2} with Delta Rule},
 url = {https://proceedings.iclr.cc/paper_files/paper/2025/file/4904fad153f6434a7bcf04465d4be2cc-Paper-Conference.pdf},
 volume = {2025},
 year = {2025}
}

@article{dao2024mamba2,
  title={Transformers are {SSMs}: Generalized models and efficient algorithms through structured state space duality},
  author={Dao, Tri and Gu, Albert},
  journal={arXiv preprint arXiv:2405.21060},
  year={2024}
}

@article{nvidia2025nemotronnano2,
  title={{NVIDIA Nemotron Nano 2}: An accurate and efficient hybrid {Mamba-Transformer} reasoning model},
  author={Basant, Aarti and Khairnar, Abhijit and Paithankar, Abhijit and Khattar, Abhinav and Renduchintala, Adithya and Malte, Aditya and Bercovich, Akhiad and Hazare, Akshay and Rico, Alejandra and Ficek, Aleksander and others},
  journal={arXiv preprint arXiv:2508.14444},
  year={2025}
}

@article{falcon2026h1r,
title={{Falcon-H1R}: Pushing the reasoning frontiers with a hybrid model for efficient test-time scaling},
  author={Chaabane, Iheb and Khanna, Puneesh and Mohmad, Suhail and Frikha, Slim and Hu, Shi and Abubaker, Abdalgader and Alami, Reda and Lubinets, Mikhail and Seddik, Mohamed El Amine and Hacid, Hakim and others},
  journal={arXiv preprint arXiv:2601.02346},
  year={2026}
}

@article{chen2024overthinking,
title={Do not think that much for 2+ 3=? on the overthinking of {o1}-like {LLMs}},
  author={Chen, Xingyu and Xu, Jiahao and Liang, Tian and He, Zhiwei and Pang, Jianhui and Yu, Dian and Song, Linfeng and Liu, Qiuzhi and Zhou, Mengfei and Zhang, Zhuosheng and others},
  journal={arXiv preprint arXiv:2412.21187},
  year={2024}
}

@InProceedings{pipis2025loop,
  title = 	 {Wait, Wait, Wait... Why Do Reasoning Models Loop?},
  author =       {Pipis, Charilaos and Garg, Shivam and Kontonis, Vasilis and Shrivastava, Vaishnavi and Krishnamurthy, Akshay and Papailiopoulos, Dimitris},
  booktitle = 	 {Proceedings of the 43rd International Conference on Machine Learning},
  pages = 	 {98934--98964},
  year = 	 {2026},
  editor = 	 {Zhang, Tong and Dudik, Miroslav and Jaggi, Martin and Agarwal, Alekh and Li, Sharon and Schuurmans, Dale and Zhu, Jerry and Berkenkamp, Felix and Dong, Hanze and Bietti, Alberto},
  volume = 	 {306},
  series = 	 {Proceedings of Machine Learning Research},
  month = 	 {06--11 Jul},
  publisher =    {PMLR},
  url = 	 {https://proceedings.mlr.press/v306/pipis26a.html}
}

@inproceedings{wang2025nowait,
    title = "Wait, We Don{'}t Need to ``Wait''! Removing Thinking Tokens Improves Reasoning Efficiency",
    author = "Wang, Chenlong  and
      Feng, Yuanning  and
      Chen, Dongping  and
      Chu, Zhaoyang  and
      Krishna, Ranjay  and
      Zhou, Tianyi",
    editor = "Christodoulopoulos, Christos  and
      Chakraborty, Tanmoy  and
      Rose, Carolyn  and
      Peng, Violet",
    booktitle = "Findings of the {A}ssociation for {C}omputational {L}inguistics: {EMNLP} 2025",
    month = nov,
    year = "2025",
    address = "Suzhou, China",
    publisher = "Association for Computational Linguistics",
    url = "https://aclanthology.org/2025.findings-emnlp.394/",
    doi = "10.18653/v1/2025.findings-emnlp.394",
    pages = "7459--7482",
    ISBN = "979-8-89176-335-7"
}

@article{chen2025seal,
title={{SEAL}: Steerable reasoning calibration of large language models for free},
  author={Chen, Runjin and Zhang, Zhenyu and Hong, Junyuan and Kundu, Souvik and Wang, Zhangyang},
  journal={arXiv preprint arXiv:2504.07986},
  year={2025}
}

@inproceedings{azizi2025asc,
    title = "Activation Steering for Chain-of-Thought Compression",
    author = "Azizi, Seyedarmin  and
      Potraghloo, Erfan Baghaei  and
      Kundu, Souvik  and
      Pedram, Massoud",
    editor = "Liakata, Maria  and
      Moreira, Viviane P.  and
      Zhang, Jiajun  and
      Jurgens, David",
    booktitle = "Findings of the {A}ssociation for {C}omputational {L}inguistics: {ACL} 2026",
    month = jul,
    year = "2026",
    address = "San Diego, California, United States",
    publisher = "Association for Computational Linguistics",
    url = "https://aclanthology.org/2026.findings-acl.1828/",
    doi = "10.18653/v1/2026.findings-acl.1828",
    pages = "36676--36687",
    ISBN = "979-8-89176-395-1"
}

@inproceedings{sun2025thinkedit,
    title = "{T}hink{E}dit: Interpretable Weight Editing to Mitigate Overly Short Thinking in Reasoning Models",
    author = "Sun, Chung-En  and
      Yan, Ge  and
      Weng, Tsui-Wei",
    editor = "Christodoulopoulos, Christos  and
      Chakraborty, Tanmoy  and
      Rose, Carolyn  and
      Peng, Violet",
    booktitle = "Proceedings of the 2025 Conference on Empirical Methods in Natural Language Processing",
    month = nov,
    year = "2025",
    address = "Suzhou, China",
    publisher = "Association for Computational Linguistics",
    url = "https://aclanthology.org/2025.emnlp-main.861/",
    doi = "10.18653/v1/2025.emnlp-main.861",
    pages = "17001--17025",
    ISBN = "979-8-89176-332-6"
}

@inproceedings{azizi2025mambaextend,
 author = {Azizi, Seyedarmin and Kundu, Souvik and Sadeghi, Mohammad and Pedram, Massoud},
 booktitle = {International Conference on Learning Representations},
 editor = {Y. Yue and A. Garg and N. Peng and F. Sha and R. Yu},
 pages = {97075--97090},
title = {{MambaExtend}: A Training-Free Approach to Improve Long Context Extension of {Mamba}},
 url = {https://proceedings.iclr.cc/paper_files/paper/2025/file/f1922bd718528ac3eab114eabbbfa7a0-Paper-Conference.pdf},
 volume = {2025},
 year = {2025}
}

@article{lu2025modulation,
  title={{Mamba Modulation}: On the Length Generalization of {Mamba}},
  author={Lu, Peng and Huang, Jerry and Zeng, Qiuhao and Wang, Xinyu and Chen, Boxing and Langlais, Philippe and Cui, Yufei},
  journal={arXiv preprint arXiv:2509.19633},
  year={2025}
}

@article{chen2024stuffed,
title={Stuffed {Mamba}: Oversized states lead to the inability to forget},
  author={Chen, Yingfa and Zhang, Xinrong and Hu, Shengding and Han, Xu and Liu, Zhiyuan and Sun, Maosong},
  journal={arXiv preprint arXiv:2410.07145},
  year={2024}
}

@inproceedings{afendulev2026recalls,
    title = "What Attention Recalls and Recurrence Controls in Hybrid Language Models",
    author = "Afendulev, Kirill  and
      Dontsov, Alexey  and
      Tutubalina, Elena  and
      Korznikov, Anton",
    editor = "Lee, Hung-Yi  and
      Dev, Sunipa  and
      Li, Jessy  and
      Elliott, Desmond",
    booktitle = "Findings of the {A}ssociation for {C}omputational {L}inguistics: {EMNLP} 2026",
    month = oct,
    year = "2026",
    address = "Budapest, Hungary",
    publisher = "Association for Computational Linguistics",
    url = "https://aclanthology.org/2026.findings-emnlp.2061/",
    doi = "10.18653/2026.findings-emnlp.2061",
    pages = "43888--43900",
    ISBN = "979-8-89176-462-0"
}

@article{cobbe2021gsm8k,
  title={Training verifiers to solve math word problems},
  author={Cobbe, Karl and Kosaraju, Vineet and Bavarian, Mohammad and Chen, Mark and Jun, Heewoo and Kaiser, Lukasz and Plappert, Matthias and Tworek, Jerry and Hilton, Jacob and Nakano, Reiichiro and others},
  journal={arXiv preprint arXiv:2110.14168},
  year={2021}
}

@article{hendrycks2021math,
    title={Measuring mathematical problem solving with the {MATH} dataset},
  author={Hendrycks, Dan and Burns, Collin and Kadavath, Saurav and Arora, Akul and Basart, Steven and Tang, Eric and Song, Dawn and Steinhardt, Jacob},
  journal={arXiv preprint arXiv:2103.03874},
  year={2021}
}

@inproceedings{lightman2024verify,
 author = {Lightman, Hunter and Kosaraju, Vineet and Burda, Yuri and Edwards, Harrison and Baker, Bowen and Lee, Teddy and Leike, Jan and Schulman, John  and Sutskever, Ilya and Cobbe, Karl},
 booktitle = {International Conference on Learning Representations},
 editor = {B. Kim and Y. Yue and S. Chaudhuri and K. Fragkiadaki and M. Khan and Y. Sun},
 pages = {39578--39601},
 title = {Let\textquotesingle s Verify Step by Step},
 url = {https://proceedings.iclr.cc/paper_files/paper/2024/file/aca97732e30bcf1303bc22ac3924fd16-Paper-Conference.pdf},
 volume = {2024},
 year = {2024}
}

@article{falcon2025h1,
  title={{Falcon-H1}: A family of hybrid-head language models redefining efficiency and performance},
  author={Zuo, Jingwei and Velikanov, Maksim and Chahed, Ilyas and Belkada, Younes and Rhayem, Dhia Eddine and Kunsch, Guillaume and Hacid, Hakim and Yous, Hamza and Farhat, Brahim and Khadraoui, Ibrahim and others},
  journal={arXiv preprint arXiv:2507.22448},
  year={2025}
}

@article{rawat2026primitives,
  title={Reasoning Primitives in Hybrid and Non-Hybrid {LLMs}: Do Architectural Differences Yield Advantages in State-Tracking and Recall?},
  author={Rawat, Shivam and Flek, Lucie and Mai, Florian and Corr{\^e}a, Nicholas Kluge},
  journal={arXiv preprint arXiv:2604.21454},
  year={2026}
}

@article{steifer2026scratchpads,
  title={Provably Shorter Scratchpads in Hybrid {DeltaNet-Attention} Decoders},
  author={Steifer, Tomasz},
  journal={arXiv preprint arXiv:2605.16640},
  year={2026}
}

@inproceedings{kwon2023efficient,
author = {Kwon, Woosuk and Li, Zhuohan and Zhuang, Siyuan and Sheng, Ying and Zheng, Lianmin and Yu, Cody Hao and Gonzalez, Joseph and Zhang, Hao and Stoica, Ion},
title = {Efficient Memory Management for Large Language Model Serving with {PagedAttention}},
year = {2023},
isbn = {9798400702297},
publisher = {Association for Computing Machinery},
address = {New York, NY, USA},
url = {https://doi.org/10.1145/3600006.3613165},
doi = {10.1145/3600006.3613165},
booktitle = {Proceedings of the 29th Symposium on Operating Systems Principles},
pages = {611–626},
numpages = {16},
location = {Koblenz, Germany},
series = {SOSP '23}
}

@misc{qwen35,
    title  = {{Qwen3.5}: Towards Native Multimodal Agents},
    author = {{Qwen Team}},
    month  = {February},
    year   = {2026},
    url    = {https://qwen.ai/blog?id=qwen3.5}
}

@article{weng2026rounding,
  title={Lost in Rounding: bf16 Recurrent State Makes Hybrid Reasoning Models Think Longer},
  author={Weng, Yuchuan and Chen, Xin and Liu, Chi},
  journal={arXiv preprint arXiv:2611.12127},
  year={2026}
}

@inproceedings{wang2024mathvision,
 author = {Wang, Ke and Pan, Junting and Shi, Weikang and Lu, Zimu and Ren, Houxing and Zhou, Aojun and Zhan, Mingjie and Li, Hongsheng},
 booktitle = {Advances in Neural Information Processing Systems},
 doi = {10.52202/079017-3014},
 editor = {A. Globerson and L. Mackey and D. Belgrave and A. Fan and U. Paquet and J. Tomczak and C. Zhang},
 pages = {95095--95169},
 publisher = {Curran Associates, Inc.},
 title = {Measuring Multimodal Mathematical Reasoning with {MATH-Vision} Dataset},
 url = {https://proceedings.neurips.cc/paper_files/paper/2024/file/ad0edc7d5fa1a783f063646968b7315b-Paper-Datasets_and_Benchmarks_Track.pdf},
 volume = {37},
 year = {2024}
}

@inproceedings{ye2025longmamba,
 author = {Ye, Zhifan and Xia, Kejing and Fu, Yonggan and Dong, Xin and Hong, Jihoon and Yuan, Xiangchi and Diao, Shizhe and Kautz, Jan and Molchanov, Pavlo and Lin, Yingyan Celine},
 booktitle = {International Conference on Learning Representations},
 editor = {Y. Yue and A. Garg and N. Peng and F. Sha and R. Yu},
 pages = {68486--68504},
 title = {{LongMamba}: Enhancing {Mamba}\textquotesingle s Long-Context Capabilities via Training-Free Receptive Field Enlargement},
 url = {https://proceedings.iclr.cc/paper_files/paper/2025/file/ab5d50d269e52f8eed497062311ff173-Paper-Conference.pdf},
 volume = {2025},
 year = {2025}
}

@inproceedings{benkish2025decimamba,
 author = {Ben-Kish, Assaf and Zimerman, Itamar and Abu-Hussein, Shady and Cohen, Nadav and Globerson, Amir and Wolf, Lior and Giryes, Raja},
 booktitle = {International Conference on Learning Representations},
 editor = {Y. Yue and A. Garg and N. Peng and F. Sha and R. Yu},
 pages = {101148--101170},
 title = {{DeciMamba}: Exploring the Length Extrapolation Potential of {Mamba}},
 url = {https://proceedings.iclr.cc/paper_files/paper/2025/file/fae0a4535196bf7715c1aef2ccfe283f-Paper-Conference.pdf},
 volume = {2025},
 year = {2025}
}

@article{tallec2018warp,
  title={Can recurrent neural networks warp time?},
  author={Tallec, Corentin and Ollivier, Yann},
  journal={arXiv preprint arXiv:1804.11188},
  year={2018}
}

@article{mahto2021timescales,
  title={Multi-timescale representation learning in {LSTM} language models},
  author={Mahto, Shivangi and Vo, Vy A and Turek, Javier S and Huth, Alexander G},
  journal={arXiv preprint arXiv:2009.12727},
  year={2020}
}

@InProceedings{orvieto2023lru,
  title = 	 {Resurrecting Recurrent Neural Networks for Long Sequences},
  author =       {Orvieto, Antonio and Smith, Samuel L and Gu, Albert and Fernando, Anushan and Gulcehre, Caglar and Pascanu, Razvan and De, Soham},
  booktitle = 	 {Proceedings of the 40th International Conference on Machine Learning},
  pages = 	 {26670--26698},
  year = 	 {2023},
  editor = 	 {Krause, Andreas and Brunskill, Emma and Cho, Kyunghyun and Engelhardt, Barbara and Sabato, Sivan and Scarlett, Jonathan},
  volume = 	 {202},
  series = 	 {Proceedings of Machine Learning Research},
  month = 	 {23--29 Jul},
  publisher =    {PMLR},
  url = 	 {https://proceedings.mlr.press/v202/orvieto23a.html}
}

@InProceedings{schlag2021fwp,
  title = 	 {Linear Transformers Are Secretly Fast Weight Programmers},
  author =       {Schlag, Imanol and Irie, Kazuki and Schmidhuber, J{\"u}rgen},
  booktitle = 	 {Proceedings of the 38th International Conference on Machine Learning},
  pages = 	 {9355--9366},
  year = 	 {2021},
  editor = 	 {Meila, Marina and Zhang, Tong},
  volume = 	 {139},
  series = 	 {Proceedings of Machine Learning Research},
  month = 	 {18--24 Jul},
  publisher =    {PMLR},
  url = 	 {https://proceedings.mlr.press/v139/schlag21a.html}
}

@inproceedings{yang2024deltanet,
 author = {Yang, Songlin and Wang, Bailin and Zhang, Yu and Shen, Yikang and Kim, Yoon},
 booktitle = {Advances in Neural Information Processing Systems},
 doi = {10.52202/079017-3668},
 editor = {A. Globerson and L. Mackey and D. Belgrave and A. Fan and U. Paquet and J. Tomczak and C. Zhang},
 pages = {115491--115522},
 publisher = {Curran Associates, Inc.},
 title = {Parallelizing Linear Transformers with the Delta Rule over Sequence Length},
 url = {https://proceedings.neurips.cc/paper_files/paper/2024/file/d13a3eae72366e61dfdc7eea82eeb685-Paper-Conference.pdf},
 volume = {37},
 year = {2024}
}

@inproceedings{yang2024gla,
author = {Yang, Songlin and Wang, Bailin and Shen, Yikang and Panda, Rameswar and Kim, Yoon},
title = {Gated linear attention transformers with hardware-efficient training},
year = {2024},
publisher = {JMLR.org},
booktitle = {Proceedings of the 41st International Conference on Machine Learning},
articleno = {2333},
numpages = {23},
location = {Vienna, Austria},
series = {ICML'24}
}

@inproceedings{zhang2019rmsnorm,
 author = {Zhang, Biao and Sennrich, Rico},
 booktitle = {Advances in Neural Information Processing Systems},
 editor = {H. Wallach and H. Larochelle and A. Beygelzimer and F. d\textquotesingle Alch\'{e}-Buc and E. Fox and R. Garnett},
 pages = {},
 publisher = {Curran Associates, Inc.},
 title = {Root Mean Square Layer Normalization},
 url = {https://proceedings.neurips.cc/paper_files/paper/2019/file/1e8a19426224ca89e83cef47f1e7f53b-Paper.pdf},
 volume = {32},
 year = {2019}
}

@inproceedings{ali2025hidden,
    title = "The Hidden Attention of {Mamba} Models",
    author = "Ali, Ameen Ali  and
      Zimerman, Itamar  and
      Wolf, Lior",
    editor = "Che, Wanxiang  and
      Nabende, Joyce  and
      Shutova, Ekaterina  and
      Pilehvar, Mohammad Taher",
    booktitle = "Proceedings of the 63rd Annual Meeting of the {A}ssociation for {C}omputational {L}inguistics (Volume 1: Long Papers)",
    month = jul,
    year = "2025",
    address = "Vienna, Austria",
    publisher = "Association for Computational Linguistics",
    url = "https://aclanthology.org/2025.acl-long.76/",
    doi = "10.18653/v1/2025.acl-long.76",
    pages = "1516--1534",
    ISBN = "979-8-89176-251-0"
}

@inproceedings{wolf2020transformers,
    title = "Transformers: State-of-the-Art Natural Language Processing",
    author = "Wolf, Thomas  and
      Debut, Lysandre  and
      Sanh, Victor  and
      Chaumond, Julien  and
      Delangue, Clement  and
      Moi, Anthony  and
      Cistac, Pierric  and
      Rault, Tim  and
      Louf, Remi  and
      Funtowicz, Morgan  and
      Davison, Joe  and
      Shleifer, Sam  and
      von Platen, Patrick  and
      Ma, Clara  and
      Jernite, Yacine  and
      Plu, Julien  and
      Xu, Canwen  and
      Le Scao, Teven  and
      Gugger, Sylvain  and
      Drame, Mariama  and
      Lhoest, Quentin  and
      Rush, Alexander",
    editor = "Liu, Qun  and
      Schlangen, David",
    booktitle = "Proceedings of the 2020 Conference on Empirical Methods in Natural Language Processing: System Demonstrations",
    month = oct,
    year = "2020",
    address = "Online",
    publisher = "Association for Computational Linguistics",
    url = "https://aclanthology.org/2020.emnlp-demos.6/",
    doi = "10.18653/v1/2020.emnlp-demos.6",
    pages = "38--45"
}

@inbook{mia2026adaptive,
author = {Mia, Md Jueal and Amini, M. Hadi},
title = {Don't Overthink, Don't Underthink: Toward Adaptive Reasoning in Agentic AI},
year = {2026},
isbn = {9798400726385},
publisher = {Association for Computing Machinery},
address = {New York, NY, USA},
url = {https://doi.org/10.1145/3806096.3844827},
booktitle = {Proceedings of the ACM AI Leadership Summit 2026},
articleno = {28},
numpages = {5}
}

\appendix
\section{Experimental Setup and Reproducibility}
\label{app:repro}

\paragraph{Models and edit.}
We use the Hugging Face checkpoints \texttt{Qwen/Qwen3.5-2B},
\texttt{Qwen/Qwen3.5-4B} and \texttt{Qwen/Qwen3.5-9B}. For each $\kappa$ we
write an overlay checkpoint that links the original weight shards and replaces
each recurrent layer's \texttt{A\_log} tensor with $A_{\log}+\ln\kappa$; the
multi-token-prediction head of Qwen3.5, which generation does not use, is left
unchanged. The tensors are stored in float32, the addition is done in float32,
and vLLM's Gated DeltaNet layer holds \texttt{A\_log} in float32, so the edit
adds no rounding beyond float32 arithmetic. The edit reads and rewrites only
these tensors from the safetensors files, without loading the model or using a
GPU. The base and NoWait settings use the original checkpoint. Each model
alternates three Gated DeltaNet layers with one attention layer; the 2B model
has 18 Gated DeltaNet layers with 288 forget-gate heads in total, and the 4B
and 9B models have 24 layers with 768 heads each. Before any generation, the
edit script checks that it finds exactly these layers and heads and that the
end-of-thinking tag \texttt{</think>} is in the tokenizer, and the identity
check below tests that the serving engine reads the edited tensors. A model
that failed any of these checks would get no $\kappa$ settings; all three
passed.

\paragraph{Identity check.}
In the evaluation configuration itself (vLLM, bfloat16, with the recurrent
state kept in float32 as each model's configuration specifies; maximum model
length 4,096, greedy, one scored token per position), we score 179 tokens of a fixed input, a short word problem in
the chat template followed by a fixed reasoning trace, with each checkpoint.
The $\kappa=1$ overlay holds the original \texttt{A\_log} values in float32,
so its exact agreement with the base log-probabilities (largest absolute
difference $0.0$ for every model; tolerance $10^{-3}$) tests the overlay
loading path rather than the edit. The $\kappa=2$ overlay changes them by a
mean absolute 0.151, 0.188 and 0.167 nats per token for the 2B, 4B and 9B
models; the check requires more than $10^{-4}$ and more than ten times the
$\kappa=1$ difference, which shows that the engine reads the edited tensors.
vLLM computes each gate from \texttt{A\_log} and the step in fused kernels;
we did not read the served gates back from these kernels.

\paragraph{Prompt and decoding.}
Each problem is followed by \emph{``Please reason step by step, and put your
final answer within \textbackslash boxed\{\}.''} and formatted with the chat
template, with thinking on and no system prompt. All settings use the model card's thinking-mode sampling parameters (temperature 1.0, top-$p$ 0.95, top-$k$
20, presence penalty 1.5) under a budget of 34,000 generated tokens, in vLLM
with bfloat16. Each question gets one response, with a seed derived from the
question, so every setting starts from the same random state
(Table~\ref{tab:hparams}).

\paragraph{Generation.}
All runs use vLLM 0.33.0 \citep{kwon2023efficient} with PyTorch 2.14.0 and
Transformers 5.20.2, in bfloat16, on NVIDIA H200 NVL GPUs (141\,GB HBM3e)
and NVIDIA A100 PCIe GPUs (40\,GB HBM2). Each worker runs on one GPU and
takes jobs from a shared queue, each covering one model, one setting, one data
set and one shard of the questions (in the budget runs, one model, setting and shard at every budget and on every data set), so no GPU is tied to one model.
For the text benchmarks, three H200 GPUs generated the $\kappa$ sweeps, the
MATH-500 comparison and the budget runs of Appendix~\ref{app:budgets}, and two
A100 GPUs generated some of the budget runs. For MATH-Vision, four H200 GPUs
generated the base, NoWait and $\kappa=2$ settings of the 9B model and the NoWait setting of the 4B model, and eight A100 GPUs generated all settings of the 2B model
and the base and $\kappa=2$ settings of the 4B model, 2,736 responses in all
(three models, three settings, 304 questions). Every worker uses the seeds and
sampling settings listed in Table~\ref{tab:hparams}. Requests are sent in
chunks of 512 (1,024 in the budget runs) and finished answers are written
after each chunk, so an interrupted run resumes without repeating them. Before
analysis we checked that each setting answered every question exactly once.

\paragraph{NoWait.}
We follow Section~3.1 of \citet{wang2025nowait}. Its 17 reflection keywords
(\emph{wait, alternatively, hmm, but, however, alternative, another, check,
double-check, oh, maybe, verify, other, again, now, ah, any}) are mapped to
every vocabulary token whose text contains one of them, and each such token
receives a logit bias of $-100$ throughout generation. The original work
removes unrelated matches (such as \emph{Ohio} for \emph{oh}) by hand without
publishing the result, so we use a fixed rule instead: a match is kept when
the keyword starts a word and is followed only by an inflection (\emph{-s,
-es, -ed, -d, -ing}) or, for \emph{hmm}, \emph{oh} and \emph{ah}, by more of
its last letter. For Qwen3.5 this bans 180 of the 1,586 matching tokens.

\paragraph{Data.}
MATH-Dev comes from the test split of \texttt{EleutherAI/hendrycks\_math}
(5,000 problems). A problem was removed if its normalized text (lower case,
with LaTeX markup and Asymptote figure code removed) equals, or shares any
13-word sequence with, a problem of MATH-500, the GSM8K test split or AIME 2024
or 2025. This removed the 500 MATH-500 problems and 67 near-duplicates of
them; of the problems at levels 3--5, three more shared a 13-word sequence
with an AIME problem, leaving 3,254. From these we sampled 200 with seed 0,
allocating them to subjects in proportion to each subject's size; the sample
has 65, 59 and 76 problems at levels 3, 4 and 5. Their identifiers are listed
in our code release. MATH-500 covers all five difficulty levels (43, 90,
105, 128 and 134 problems) and played no part in choosing $\kappa$ or any
other setting. GSM8K (the full test split of 1,319 problems) comes from
\texttt{openai/gsm8k}, AIME 2024 from \texttt{HuggingFaceH4/aime\_2024}, AIME
2025 from \texttt{MathArena/aime\_2025} (60 problems in all) and MATH-500
from \texttt{HuggingFaceH4/MATH-500}. MATH-Vision testmini
\citep{wang2024mathvision} has 304 competition problems posed with a diagram
or figure, at five difficulty levels, 190 of them multiple-choice and 114
free-form.

\paragraph{Grading and counting.}
We take the last \texttt{\textbackslash boxed\{\}} expression after the
end-of-thinking tag. A GSM8K answer is correct when it equals the reference
numerically, also when the box holds a single number with a unit or currency
sign; an AIME answer when it equals the reference numerically or
\texttt{math-verify} accepts it; a MATH or MATH-500 answer when the normalized
strings or the numeric values agree or \texttt{math-verify} accepts it. A
free-form MATH-Vision answer is graded as a MATH answer, and a
multiple-choice one by its option letter or by the text of exactly one
option. A
response without an end-of-thinking tag is a budget hit and counts as wrong,
even if its thinking contains a boxed answer. Thinking tokens are the tokens
before the end-of-thinking tag, or all generated tokens for a budget hit;
total tokens are thinking plus answer tokens, and the budget also counts the
end-of-thinking tag and the end-of-turn token. Re-check phrases are matched
case-insensitively from the start of a word, so \emph{waiting} also counts;
\emph{actually} counts only when followed by a comma, and \emph{double check}
counts with or without the hyphen. A loop is a 64-token window of the
thinking that occurs at least four times, overlapping occurrences included.
Token changes are computed as the ratio of summed tokens minus one, and the
bootstrap uses resampling seed 0. Accuracy among finished responses is the
number of correct responses divided by the number of responses that close
their thinking within the budget.

\paragraph{Measures and statistics.}
Besides accuracy, mean thinking tokens, the share of responses that reach
the budget and accuracy among finished responses, we count looping traces
(the loop rule above, fixed before any run) and re-check phrases
(\emph{wait, hmm, let me check, double-check, let me verify, actually,
alternatively}) per 1,000 thinking tokens. Confidence intervals are 95\%
paired bootstrap intervals over questions, with 10,000 resamples (2,000 per
MATH level), and are not corrected for multiple comparisons.

\begin{table}
  \centering
  \begin{tabular}{@{}ll@{}}
    \toprule
    Setting & Value \\
    \midrule
    Temp. / top-$p$ / top-$k$ & 1.0 / 0.95 / 20 \\
    Presence penalty & 1.5 \\
    System prompt & none \\
    Generation budget & 34,000 tokens \\
    Max model length & 36,048 \\
    Precision & bfloat16 (vLLM) \\
    GPUs & H200 NVL, A100 PCIe \\
    Recurrent state & float32 \\
    Samples per question & 1 \\
    Seed & per question (global 0) \\
    $\kappa$ values & 0.5 to 3 in steps of 0.25 \\
    NoWait & 180 tokens banned \\
    Loop detector & 64-token span $\geq 4$ times \\
    Bootstrap resamples & 10,000 (2,000 per level) \\
    \bottomrule
  \end{tabular}
  \caption{Values shared by all models and settings. The maximum model length
  is the budget plus 2,048; the budget runs (Appendix~\ref{app:budgets}) use
  18,432 and the MATH-Vision runs 52,432. NoWait gives each banned token a
  logit bias of $-100$.}
  \label{tab:hparams}
\end{table}

\section{Additional Results}
\label{app:full}

\begin{table*}[t]
  \centering
  \resizebox{\linewidth}{!}{%
  \begin{tabular}{@{}lccccccc@{}}
    \toprule
    Model & Acc. & $\Delta$Acc. & $\Delta$Th. & Bud. & Gain & Loss & FAcc. \\
          & B$\to$2 & [CI] & [CI] & B$\to$2 & u/w & u/w & B$\to$2 \\
    \midrule
    Qwen3.5-2B & 66.2$\to$76.2 & \cellcolor{bestcell}$+10.0$ [6.4, 13.6] & \cellcolor{bestcell}$-11.7$ [$-14.5$, $-8.8$] & 32.2$\to$19.0 & 65/4 & 11/8 & 97.6$\to$94.1 \\
    Qwen3.5-4B & 94.8$\to$96.4 & $+1.6$ [$-0.4$, 3.6] & \cellcolor{bestcell}$-17.2$ [$-20.9$, $-13.5$] & 4.4$\to$2.2 & 14/3 & 4/5 & 99.2$\to$98.6 \\
    Qwen3.5-9B & 96.4$\to$98.0 & $+1.6$ [$-0.2$, 3.4] & \cellcolor{bestcell}$-18.6$ [$-22.4$, $-14.7$] & 2.8$\to$1.2 & 12/2 & 3/3 & 99.2$\to$99.2 \\
    \bottomrule
  \end{tabular}%
  }
  \caption{Held-out test on MATH-500 (levels 1--5), with $\kappa=2$ fixed on MATH-Dev beforehand. B$\to$2: base $\to$ $\kappa=2$. Acc.: accuracy (\%); $\Delta$Acc.: its change (pp); $\Delta$Th.: change in thinking tokens (\%); [CI]: 95\% bootstrap interval; Bud.: responses reaching the budget (\%); Gain: questions the edit answers correctly that the base model left unfinished (u) or answered wrongly (w); Loss: the reverse; FAcc.: accuracy among finished responses (\%). Shaded: the change favours the edit (less thinking or higher accuracy) and its 95\% interval excludes zero.}
  \label{tab:heldout}
\end{table*}

\paragraph{Held-out MATH-500 in detail.}
\label{app:heldout}
Thinking plus answer tokens fell by about as much as thinking alone
(10.9--17.8\%). The saving depends on difficulty. On levels 3--5 every interval lies below
zero, whereas on level~1 only the 2B model changed its thinking detectably
($-17.5\%$; 4B: $-11.2\%$; 9B: $-6.4\%$). As on MATH-Dev, where it saved
10.7\% on level~3 and 25--27\% on levels 4 and 5, the 9B model saved more on
harder problems, from 6.4\% on level~1 to 21.6\% on level~5. The 2B model
saved least on level~5 (6.5\%), where 57.5\% of its base responses reach the
budget. Because both settings answer the same questions, each question can be
followed from the base model to the edit (Table~\ref{tab:heldout}). For the
2B model, the edit finished 79 of the 161 questions the base model left
unfinished and answered 65 of them correctly; with 4 corrected answers and
19 losses (11 by not finishing), its net gain was 50 questions: $+54$ from
responses that now finish (65 gained, 11 lost) and $-4$ among questions
that both settings finished (4 corrected, 8 lost). The 4B and 9B
models gained 17 and 14 questions and lost 9 and 6. These transitions are
measured at a single budget.

Table~\ref{tab:main} lists accuracy, mean thinking tokens and the share of
responses that reach the budget for every model, benchmark and value of $\kappa$, including $\kappa=2.25$ to 3, which the figures omit.
\begin{table*}
  \centering\footnotesize\setlength{\tabcolsep}{4pt}
  \begin{tabular}{@{}lccccccccc@{}}
    \toprule
    & \multicolumn{3}{c}{MATH-Dev (L3--5)} & \multicolumn{3}{c}{GSM8K} & \multicolumn{3}{c}{AIME 24+25} \\
    \cmidrule(lr){2-4}\cmidrule(lr){5-7}\cmidrule(lr){8-10}
    $\kappa$ & Acc. & Think. & Bud. & Acc. & Think. & Bud. & Acc. & Think. & Bud. \\
    \midrule
    \multicolumn{10}{l}{\emph{Qwen3.5-2B}} \\
    1 (base) & 66.5 & 21.6 & 32.0 & 82.7 & 13.0 & 13.4 & 13.3 & 32.3 & 86.7 \\
    0.5 & 25.0 & 28.4 ($+31.7$) & 73.0 & 52.1 & 21.0 ($+61.2$) & 44.5 & 0.0 & 34.0 ($+5.1$) & 100.0 \\
    0.75 & 47.0 & 24.0 ($+11.1$) & 51.0 & 73.5 & 15.5 ($+19.4$) & 22.3 & 5.0 & 33.4 ($+3.1$) & 93.3 \\
    1.25 & 66.5 & 20.9 ($-3.1$) & 25.0 & 84.2 & 11.7 ($-10.1$) & 9.2 & 15.0 & 32.6 ($+0.9$) & 85.0 \\
    1.5 & 72.5 & 19.5 ($-9.6$) & 22.0 & 86.0 & 11.0 ($-15.3$) & 7.5 & 13.3 & 31.7 ($-2.0$) & 85.0 \\
    1.75 & 74.5 & 19.3 ($-10.5$) & 22.5 & \cellcolor{bestcell}86.6 & 10.7 ($-17.5$) & 6.4 & 15.0 & 31.9 ($-1.3$) & 80.0 \\
    2 & \cellcolor{bestcell}76.0 & 18.6 ($-13.7$) & 20.0 & 85.4 & 10.6 ($-18.6$) & 4.9 & 18.3 & 31.2 ($-3.4$) & 75.0 \\
    2.25 & 73.0 & 19.4 ($-10.0$) & 20.0 & 83.4 & 10.8 ($-17.1$) & 6.4 & \cellcolor{bestcell}20.0 & 30.5 ($-5.8$) & 65.0 \\
    2.5 & 72.5 & \cellcolor{bestcell}18.5 ($-14.3$) & 17.0 & 83.5 & 10.6 ($-18.4$) & 6.7 & 16.7 & 29.9 ($-7.6$) & 61.7 \\
    2.75 & 63.5 & 19.1 ($-11.7$) & 23.0 & 81.3 & \cellcolor{bestcell}10.5 ($-19.3$) & 6.4 & 11.7 & 29.5 ($-8.9$) & 60.0 \\
    3 & 59.0 & 19.4 ($-10.2$) & 20.0 & 77.3 & 10.7 ($-17.4$) & 8.7 & 10.0 & \cellcolor{bestcell}28.6 ($-11.5$) & 56.7 \\
    \midrule
    \multicolumn{10}{l}{\emph{Qwen3.5-4B}} \\
    1 (base) & 95.0 & 9.7 & 3.0 & \cellcolor{bestcell}95.7 & 4.1 & 0.8 & 51.7 & 25.6 & 46.7 \\
    0.5 & 81.5 & 15.1 ($+55.4$) & 17.0 & 93.9 & 6.1 ($+50.8$) & 2.9 & 21.7 & 31.1 ($+21.3$) & 73.3 \\
    0.75 & 91.0 & 11.5 ($+18.2$) & 7.0 & 94.7 & 4.7 ($+14.8$) & 1.3 & 48.3 & 27.2 ($+6.1$) & 51.7 \\
    1.25 & 96.0 & 8.9 ($-8.3$) & 2.0 & 95.5 & 3.8 ($-6.3$) & 0.4 & 61.7 & 25.1 ($-1.9$) & 38.3 \\
    1.5 & 96.5 & 8.7 ($-10.3$) & 2.0 & 95.3 & 3.9 ($-4.7$) & 0.8 & 60.0 & 23.5 ($-8.5$) & 36.7 \\
    1.75 & 95.0 & 8.7 ($-10.6$) & 3.0 & 95.5 & 3.7 ($-9.7$) & 0.7 & 63.3 & 22.6 ($-11.8$) & 26.7 \\
    2 & 95.5 & 8.5 ($-12.2$) & 2.0 & 95.3 & \cellcolor{bestcell}3.6 ($-10.4$) & 0.4 & \cellcolor{bestcell}71.7 & 21.3 ($-17.0$) & 21.7 \\
    2.25 & 94.5 & 8.0 ($-17.2$) & 2.5 & 95.1 & 3.7 ($-8.3$) & 0.5 & 65.0 & 22.3 ($-13.0$) & 26.7 \\
    2.5 & \cellcolor{bestcell}97.0 & \cellcolor{bestcell}8.0 ($-18.1$) & 1.0 & 95.6 & 3.8 ($-6.1$) & 0.2 & 65.0 & 21.2 ($-17.4$) & 28.3 \\
    2.75 & 94.0 & 8.2 ($-15.8$) & 1.5 & 94.7 & 3.8 ($-6.5$) & 0.3 & 53.3 & 21.7 ($-15.4$) & 25.0 \\
    3 & \cellcolor{bestcell}97.0 & 8.1 ($-16.9$) & 0.0 & 94.8 & 4.0 ($-0.7$) & 0.7 & 55.0 & \cellcolor{bestcell}20.8 ($-18.7$) & 20.0 \\
    \midrule
    \multicolumn{10}{l}{\emph{Qwen3.5-9B}} \\
    1 (base) & 96.0 & 9.0 & 3.0 & 96.4 & 3.3 & 0.1 & 66.7 & 23.7 & 31.7 \\
    0.5 & 83.5 & 13.3 ($+48.5$) & 14.0 & 94.8 & 5.2 ($+57.6$) & 1.5 & 26.7 & 30.2 ($+27.2$) & 70.0 \\
    0.75 & 95.0 & 10.2 ($+13.1$) & 4.0 & 96.6 & 3.7 ($+13.6$) & 0.2 & 48.3 & 26.1 ($+9.9$) & 48.3 \\
    1.25 & \cellcolor{bestcell}99.5 & 8.1 ($-10.1$) & 0.0 & \cellcolor{bestcell}96.7 & 3.1 ($-6.2$) & 0.2 & \cellcolor{bestcell}75.0 & 21.6 ($-8.9$) & 25.0 \\
    1.5 & 96.5 & 7.9 ($-12.5$) & 2.0 & 96.2 & 2.9 ($-10.4$) & 0.2 & 71.7 & 20.7 ($-12.6$) & 23.3 \\
    1.75 & 98.5 & 7.2 ($-19.7$) & 0.5 & 96.4 & 2.8 ($-13.1$) & 0.1 & \cellcolor{bestcell}75.0 & 19.9 ($-16.0$) & 23.3 \\
    2 & 97.0 & 7.0 ($-22.6$) & 1.5 & 95.8 & 2.9 ($-10.4$) & 0.3 & 68.3 & 20.7 ($-12.9$) & 30.0 \\
    2.25 & 98.5 & \cellcolor{bestcell}6.8 ($-24.8$) & 0.5 & 95.6 & \cellcolor{bestcell}2.8 ($-13.9$) & 0.1 & 70.0 & 20.2 ($-14.7$) & 23.3 \\
    2.5 & 97.5 & 7.0 ($-22.1$) & 0.0 & 95.6 & 2.9 ($-12.4$) & 0.1 & \cellcolor{bestcell}75.0 & 18.2 ($-23.2$) & 8.3 \\
    2.75 & 96.0 & 6.8 ($-23.9$) & 0.5 & 96.0 & 2.8 ($-13.7$) & 0.0 & 70.0 & \cellcolor{bestcell}18.1 ($-23.7$) & 11.7 \\
    3 & 96.5 & 7.1 ($-21.4$) & 1.0 & 95.8 & 2.9 ($-10.6$) & 0.2 & 66.7 & 18.4 ($-22.6$) & 13.3 \\
    \bottomrule
  \end{tabular}
  \caption{Full results for every value of $\kappa$, the values behind Figures~\ref{fig:tokens} and~\ref{fig:acc}. $\kappa<1$ forgets more slowly than the unedited model ($\kappa=1$) and $\kappa>1$ faster; the figures show $\kappa\le 2$. Acc.: accuracy (\%). Think.: mean thinking tokens in thousands, with the change from the base model in parentheses (\%). Bud.: share of responses that reach the 34,000-token budget (\%), which count as wrong. MATH-Dev uses levels 3--5; AIME combines 2024 and 2025. Shaded: the highest accuracy and the shortest mean thinking for each model and benchmark (ties shaded together); shading marks the best observed value, not a selected $\kappa$.}
  \label{tab:main}
\end{table*}

\paragraph{Size of the change across $\kappa$.}
\label{app:monotone}
Across all seven values of $\kappa$ up to 2, mean thinking fell strictly as
$\kappa$ rose in five of the nine model--benchmark pairs; in the other four
it rose slightly between some neighbouring values of $\kappa\ge 1$, by at
most 3.1\% of the unedited length. The clearest case is the 2B model on
AIME, where most traces reach the budget at every value and, for $\kappa>1$, the change stays
between $+0.9\%$ and $-3.4\%$.

\paragraph{Values of $\kappa$ above 2.}
\label{app:above2}
The same experiments also covered $\kappa=2.25$ to 3 (Table~\ref{tab:main}).
There the mean saving over the nine model--benchmark pairs stayed at
13.9--15.5\% (13.5\% at $\kappa=2$), while the mean accuracy change fell from
$+4.4$~pp at $\kappa=2$ to $-1.3$~pp at $\kappa=3$, with detected losses for
the 2B model at $\kappa=3$ and the 9B model on GSM8K at $\kappa=2.25$.

\paragraph{Budget and completion on the development benchmarks.}
At $\kappa=2$ the share of responses that reach the budget fell in eight of
the nine settings (Table~\ref{tab:k2}), most for the 2B model on MATH-Dev
(32.0\% to 20.0\%) and GSM8K (13.4\% to 4.9\%) and for the 4B model on AIME
(46.7\% to 21.7\%), and rose slightly only for the 9B model on GSM8K (0.1\%
to 0.3\%), where almost no trace runs out. Accuracy among finished responses
did not rise along with completion: for the 2B model it fell from 97.8\% to
95.0\% on MATH-Dev and from 95.5\% to 89.7\% on GSM8K, and for the 4B and 9B
models on these two benchmarks it stayed within 1~pp of the base model. On
MATH-Dev the 2B model finished 24 more responses and answered 19 more
correctly, so its accuracy gain is smaller than its gain in completion.
Different questions finish in each setting, so these aggregates cannot show
which questions moved between unfinished, wrong and right; the held-out test
follows each question (\S\ref{sec:heldout}).

\paragraph{Difficulty levels on MATH-Dev.}
\begin{table}
  \centering
  \resizebox{\linewidth}{!}{%
  \begin{tabular}{@{}llccccc@{}}
    \toprule
    & & & \multicolumn{4}{c}{$\kappa$} \\
    \cmidrule(lr){4-7}
    Model & & Base & 1.25 & 1.5 & 1.75 & 2 \\
    \midrule
    2B L3 & Acc. & 78.5 & 81.5 & 81.5 & \cellcolor{bestcell}87.7 & 84.6 \\
          & Tok. & 16.9 & 16.5 & 16.0 & \cellcolor{bestcell}14.6 & 15.6 \\
    2B L4 & Acc. & 71.2 & 61.0 & 78.0 & 76.3 & \cellcolor{bestcell}79.7 \\
          & Tok. & 21.0 & 20.6 & 17.8 & 18.2 & \cellcolor{bestcell}17.4 \\
    2B L5 & Acc. & 52.6 & 57.9 & 60.5 & 61.8 & \cellcolor{bestcell}65.8 \\
          & Tok. & 26.0 & 25.0 & 23.8 & 24.2 & \cellcolor{bestcell}22.2 \\
    \midrule
    4B L3 & Acc. & 96.9 & \cellcolor{bestcell}98.5 & 95.4 & 95.4 & 96.9 \\
          & Tok. & 6.2 & \cellcolor{bestcell}5.8 & 6.2 & 6.2 & 6.1 \\
    4B L4 & Acc. & 98.3 & 96.6 & 98.3 & 98.3 & \cellcolor{bestcell}100.0 \\
          & Tok. & 8.5 & 8.3 & 8.1 & 7.4 & \cellcolor{bestcell}7.2 \\
    4B L5 & Acc. & 90.8 & 93.4 & \cellcolor{bestcell}96.1 & 92.1 & 90.8 \\
          & Tok. & 13.6 & 12.0 & \cellcolor{bestcell}11.3 & 11.8 & 11.6 \\
    \midrule
    9B L3 & Acc. & 98.5 & \cellcolor{bestcell}100.0 & 95.4 & \cellcolor{bestcell}100.0 & 96.9 \\
          & Tok. & 6.0 & 5.9 & 5.8 & \cellcolor{bestcell}5.0 & 5.4 \\
    9B L4 & Acc. & 96.6 & \cellcolor{bestcell}100.0 & \cellcolor{bestcell}100.0 & \cellcolor{bestcell}100.0 & \cellcolor{bestcell}100.0 \\
          & Tok. & 8.3 & 7.2 & 6.9 & 6.5 & \cellcolor{bestcell}6.1 \\
    9B L5 & Acc. & 93.4 & \cellcolor{bestcell}98.7 & 94.7 & 96.1 & 94.7 \\
          & Tok. & 12.0 & 10.6 & 10.4 & 9.6 & \cellcolor{bestcell}9.0 \\
    \bottomrule
  \end{tabular}%
  }
  \caption{MATH-Dev by difficulty level (L3, L4, L5: 65, 59 and 76 problems) for Qwen3.5-2B/4B/9B. Acc.: accuracy (\%); Tok.: mean thinking tokens (thousands). Shaded: the highest accuracy and the fewest thinking tokens in each row (ties shaded together).}
  \label{tab:levels}
\end{table}

Unedited models think longer as problems get harder, and at $\kappa=2$
every model thinks less at every level (Table~\ref{tab:levels}), although on
level 3 no interval lies below zero. For the 2B model the accuracy gain is
largest on level 5, from 52.6\% to 65.8\% ($+13.2$~pp, $[3.9,23.7]$), with
14.7\% less thinking, while the share of responses reaching the budget fell
from 46.1\% to 28.9\%. For the 4B and 9B models the saving is small on level
3 (1.4\% and 10.7\%, intervals include zero) and large on levels 4 and 5
(15.3\% and 14.8\% for the 4B model, 27.0\% and 25.3\% for the 9B model),
where accuracy did not fall (changes of 0.0 to $+3.4$~pp).

\paragraph{Loops on the development benchmarks.}
\label{app:loops}
At $\kappa=2$, looping traces rose slightly for the 2B model on all three
benchmarks (13 to 14 on MATH-Dev, 47 to 65 on GSM8K and 6 to 8 on AIME).
The 4B model went from 0 to 0, 2 to 4 and 1 to 0 on the three benchmarks,
and the 9B model from 1 to 0, 0 to 3 and 0 to 0. Faster forgetting thus does
not reliably prevent degenerate traces, whereas slower forgetting clearly
causes them in the 2B model ($\kappa=0.5$: 87, 455 and 43 looping traces)
and adds at most eight in the larger models.

\paragraph{Re-checking on the development benchmarks.}
On MATH-Dev at $\kappa=2$, re-check phrases per 1,000 thinking tokens barely
moved for any model: 7.99 to 7.40 for the 2B model, 2.82 to 3.01 for the 4B
model and 2.52 to 2.48 for the 9B model. On GSM8K and AIME the larger models
changed by at most 0.34, while the 2B model's density rose on GSM8K (5.03 to
6.19) and fell on AIME (11.94 to 10.37). The models thus maintain a similar
density of re-check phrases while their traces get shorter.

This is a lexical measure:
it does not show that the same verification steps survive, and paired traces
would have to be read to say which steps change.

\section{Visual Problems and NoWait}
\label{app:vision}

\begin{table*}[t]
  \centering\footnotesize\setlength{\tabcolsep}{3pt}
  \resizebox{\linewidth}{!}{%
  \begin{tabular}{@{}lccccccc@{}}
    \toprule
    & & \multicolumn{2}{c}{$\kappa=2$} & \multicolumn{2}{c}{NoWait} & \multicolumn{2}{c}{$\kappa=2$ vs.\ NoWait} \\
    \cmidrule(lr){3-4}\cmidrule(lr){5-6}\cmidrule(lr){7-8}
    Model & Base & $\Delta$Th. [CI] & $\Delta$Acc. [CI] & $\Delta$Th. [CI] & $\Delta$Acc. [CI] & $\Delta$Th. [CI] & $\Delta$Acc. [CI] \\
    \midrule
    Qwen3.5-2B & 44.4 & \cellcolor{bestcell}$-38.7$ [$-41.7$, $-35.5$] & $-3.9$ [$-9.9$, 2.0] & \cellcolor{bestcell}$-17.1$ [$-21.4$, $-12.6$] & $-4.3$ [$-9.5$, 1.0] & \cellcolor{bestcell}$-26.0$ [$-30.1$, $-21.6$] & $+0.3$ [$-5.6$, 6.2] \\
    Qwen3.5-4B & 64.8 & \cellcolor{bestcell}$-23.4$ [$-27.1$, $-19.4$] & $+0.3$ [$-4.3$, 4.9] & \cellcolor{bestcell}$-21.4$ [$-25.7$, $-17.0$] & $+1.6$ [$-3.3$, 6.6] & $-2.5$ [$-8.2$, 3.9] & $-1.3$ [$-6.2$, 3.6] \\
    Qwen3.5-9B & 67.8 & \cellcolor{bestcell}$-24.7$ [$-28.7$, $-20.5$] & $-0.7$ [$-5.3$, 3.9] & \cellcolor{bestcell}$-18.7$ [$-23.3$, $-13.9$] & $+2.0$ [$-2.6$, 6.6] & \cellcolor{bestcell}$-7.4$ [$-13.0$, $-1.5$] & $-2.6$ [$-7.2$, 2.0] \\
    \bottomrule
  \end{tabular}%
  }
  \caption{MATH-Vision testmini (304 problems with figures). Base: accuracy of the unedited model (\%). $\Delta$Th., $\Delta$Acc.: change in thinking tokens (\%) and accuracy (pp) against the base model, paired by question, with 95\% bootstrap intervals. $\kappa=2$ vs.\ NoWait: the edit against NoWait on the same questions (negative $\Delta$Th.: the edit thinks less). Shaded: the interval excludes zero.}
  \label{tab:vision}
\end{table*}

Table~\ref{tab:vision-levels} breaks the MATH-Vision results of
\S\ref{sec:vision} down by difficulty level and answer format. The edit
shortened thinking in every subset for every model, with every interval
below zero, and by more than NoWait in 19 of the 21 model--subset pairs.
Changes in accuracy are mostly within the noise of these small subsets; the
one detected change is the 2B model's loss on level~2 under the edit
($-15.9$~pp, 82 problems).

\begin{table*}
  \centering\footnotesize\setlength{\tabcolsep}{3pt}
  \begin{tabular}{@{}lrcccccccccccc@{}}
    \toprule
    & & \multicolumn{4}{c}{Qwen3.5-2B} & \multicolumn{4}{c}{Qwen3.5-4B} & \multicolumn{4}{c}{Qwen3.5-9B} \\
    \cmidrule(lr){3-6}\cmidrule(lr){7-10}\cmidrule(lr){11-14}
    & & \multicolumn{2}{c}{$\Delta$Th.} & \multicolumn{2}{c}{$\Delta$Acc.} & \multicolumn{2}{c}{$\Delta$Th.} & \multicolumn{2}{c}{$\Delta$Acc.} & \multicolumn{2}{c}{$\Delta$Th.} & \multicolumn{2}{c}{$\Delta$Acc.} \\
    \cmidrule(lr){3-4}\cmidrule(lr){5-6}\cmidrule(lr){7-8}\cmidrule(lr){9-10}\cmidrule(lr){11-12}\cmidrule(lr){13-14}
    Subset & $n$ & $\kappa{=}2$ & NoWait & $\kappa{=}2$ & NoWait & $\kappa{=}2$ & NoWait & $\kappa{=}2$ & NoWait & $\kappa{=}2$ & NoWait & $\kappa{=}2$ & NoWait \\
    \midrule
    Level 1 & 53 & \cellcolor{bestcell}$-45.2^{*}$ & $-21.0^{*}$ & \cellcolor{bestcell}$+7.5$ & $+1.9$ & \cellcolor{bestcell}$-31.5^{*}$ & $-24.3^{*}$ & $+1.9$ & \cellcolor{bestcell}$+3.8$ & \cellcolor{bestcell}$-30.0^{*}$ & $-22.2^{*}$ & \cellcolor{bestcell}$+7.5$ & \cellcolor{bestcell}$+7.5$ \\
    Level 2 & 82 & \cellcolor{bestcell}$-35.0^{*}$ & $-24.2^{*}$ & $-15.9^{*}$ & \cellcolor{bestcell}$-8.5$ & \cellcolor{bestcell}$-24.7^{*}$ & $-22.5^{*}$ & \cellcolor{bestcell}$+4.9$ & \cellcolor{bestcell}$+4.9$ & $-16.6^{*}$ & \cellcolor{bestcell}$-20.6^{*}$ & \cellcolor{bestcell}$+1.2$ & \cellcolor{bestcell}$+1.2$ \\
    Level 3 & 56 & \cellcolor{bestcell}$-35.4^{*}$ & $-21.1^{*}$ & \cellcolor{bestcell}$+3.6$ & $-5.4$ & $-23.2^{*}$ & \cellcolor{bestcell}$-26.5^{*}$ & $-5.4$ & \cellcolor{bestcell}$0.0$ & \cellcolor{bestcell}$-22.0^{*}$ & $-16.5^{*}$ & $-1.8$ & \cellcolor{bestcell}$+5.4$ \\
    Level 4 & 45 & \cellcolor{bestcell}$-32.1^{*}$ & $-5.6$ & $-13.3$ & \cellcolor{bestcell}$-6.7$ & \cellcolor{bestcell}$-14.2^{*}$ & $-14.0^{*}$ & \cellcolor{bestcell}$+6.7$ & $0.0$ & \cellcolor{bestcell}$-30.9^{*}$ & $-18.5^{*}$ & $-2.2$ & \cellcolor{bestcell}$+2.2$ \\
    Level 5 & 68 & \cellcolor{bestcell}$-44.0^{*}$ & $-11.5^{*}$ & \cellcolor{bestcell}$+1.5$ & $-1.5$ & \cellcolor{bestcell}$-20.9^{*}$ & $-18.5^{*}$ & $-5.9$ & \cellcolor{bestcell}$-1.5$ & \cellcolor{bestcell}$-25.1^{*}$ & $-15.6^{*}$ & $-7.4$ & \cellcolor{bestcell}$-4.4$ \\
    \midrule
    MC & 190 & \cellcolor{bestcell}$-38.7^{*}$ & $-12.5^{*}$ & \cellcolor{bestcell}$-4.7$ & $-6.3$ & \cellcolor{bestcell}$-23.9^{*}$ & $-22.6^{*}$ & $-2.1$ & \cellcolor{bestcell}$+3.2$ & \cellcolor{bestcell}$-24.5^{*}$ & $-16.5^{*}$ & $-1.1$ & \cellcolor{bestcell}$+5.8$ \\
    Free-form & 114 & \cellcolor{bestcell}$-38.6^{*}$ & $-24.7^{*}$ & $-2.6$ & \cellcolor{bestcell}$-0.9$ & \cellcolor{bestcell}$-22.4^{*}$ & $-19.0^{*}$ & \cellcolor{bestcell}$+4.4$ & $-0.9$ & \cellcolor{bestcell}$-25.0^{*}$ & $-22.7^{*}$ & \cellcolor{bestcell}$0.0$ & $-4.4$ \\
    \bottomrule
  \end{tabular}
  \caption{MATH-Vision testmini by difficulty level and answer format, $\kappa=2$ and NoWait against the base model. $\Delta$Th.: change in thinking tokens (\%); $\Delta$Acc.: change in accuracy (pp); $^{*}$: the 95\% paired bootstrap interval excludes zero. MC: multiple-choice. Shaded: in each row and model, the setting with less thinking and the setting with higher accuracy (ties shaded together).}
  \label{tab:vision-levels}
\end{table*}

\section{A Larger Token Budget}
\label{app:budget82k}

Our main budget of 34,000 tokens leaves many responses of the 2B model
unfinished (Table~\ref{tab:k2}), and part of its accuracy gain comes from
responses that now finish (Appendix~\ref{app:heldout}). In a separate run we
therefore generated the base model and the edit with a budget of 81,920
tokens, with $\kappa=2$ unchanged and the other settings of
\S\ref{sec:setup}. Table~\ref{tab:budget82k} compares $\kappa=2$ with the
base model on MATH-Dev and held-out MATH-500, and Table~\ref{tab:main82k}
gives every value of $\kappa$ from 0.5 to 2.

\begin{table*}[t]
  \centering\small
  \begin{tabular}{@{}llcccc@{}}
    \toprule
    Model & Benchmark & Acc. (\%) & $\Delta$Thinking (\%) & $\Delta$Acc. (pp) & Budget hits (\%) \\
          & & base$\to\kappa{=}2$ & [95\% CI] & [95\% CI] & base$\to\kappa{=}2$ \\
    \midrule
    2B & MATH-Dev & 80.5$\to$80.0 & \cellcolor{bestcell}$-22.3$ [$-30.5$, $-13.3$] & $-0.5$ [$-6.5$, 5.5] & 14.0$\to$9.0 \\
       & MATH-500 & 85.0$\to$84.2 & \cellcolor{bestcell}$-24.1$ [$-28.6$, $-19.2$] & $-0.8$ [$-4.2$, 2.4] & 12.0$\to$6.2 \\
    \midrule
    4B & MATH-Dev & 98.0$\to$97.0 & \cellcolor{bestcell}$-22.4$ [$-27.7$, $-16.3$] & $-1.0$ [$-3.0$, 1.0] & 0.5$\to$0.5 \\
       & MATH-500 & 99.0$\to$97.6 & \cellcolor{bestcell}$-19.6$ [$-24.2$, $-14.5$] & $-1.4$ [$-2.8$, 0.0] & 0.4$\to$0.0 \\
    \midrule
    9B & MATH-Dev & 98.0$\to$97.5 & \cellcolor{bestcell}$-21.1$ [$-27.3$, $-14.2$] & $-0.5$ [$-2.5$, 1.5] & 0.5$\to$0.5 \\
       & MATH-500 & 99.0$\to$98.8 & \cellcolor{bestcell}$-21.0$ [$-25.6$, $-15.8$] & $-0.2$ [$-1.0$, 0.4] & 0.2$\to$0.0 \\
    \bottomrule
  \end{tabular}
  \caption{The edit at $\kappa=2$ against the unedited model with a budget of 81,920 tokens, on MATH-Dev (levels 3--5) and held-out MATH-500 (levels 1--5). 2B/4B/9B: Qwen3.5-2B/4B/9B. Changes are paired by question, with 95\% bootstrap intervals. Budget hits: responses that reach the budget, which count as wrong. Shaded: the change favours the edit (less thinking) and its 95\% interval excludes zero.}
  \label{tab:budget82k}
\end{table*}

\paragraph{The saving holds at a larger budget.}
At 81,920 tokens the edit shortened thinking in all six settings, by
19.6--24.1\%, with every interval below zero, and thinking plus answer
tokens fell by 17.0--22.6\%. The saving was at least as large as at 34,000
tokens except for the 9B model on MATH-Dev (21.1\% against 22.6\%). No
change in accuracy was detected in any of the six settings (changes of
$-1.4$ to $-0.2$~pp), although for the 4B model on MATH-500 the interval
reaches zero ($-1.4$~pp [$-2.8$, $0.0$]). With more room to finish, the base
models reach the budget far less often (at most 14.0\%, against 32.0\% for
the 2B model on MATH-Dev at 34,000 tokens), which leaves little room for the
completion gains of the main run (Appendix~\ref{app:heldout}); accuracy
instead stays close to that of the unedited model, while the saving in
thinking persists.

\paragraph{Every value of $\kappa$.}
As at 34,000 tokens, every $\kappa<1$ lengthened thinking, with all 12
intervals above zero, and every $\kappa>1$ shortened it, with the interval
below zero in 23 of the 24 settings (Table~\ref{tab:main82k}); the
exception is the 4B model on MATH-Dev at $\kappa=1.25$ ($-7.7\%$). Slower
forgetting again lowered accuracy most for the 2B model, from 80.5\% to
30.0\% on MATH-Dev at $\kappa=0.5$.

\begin{table}[t]
  \centering\footnotesize\setlength{\tabcolsep}{3pt}
  \resizebox{\linewidth}{!}{%
  \begin{tabular}{@{}lcccccc@{}}
    \toprule
    & \multicolumn{3}{c}{MATH-Dev (L3--5)} & \multicolumn{3}{c}{MATH-500 (L1--5)} \\
    \cmidrule(lr){2-4}\cmidrule(lr){5-7}
    $\kappa$ & Acc. & Think. & Bud. & Acc. & Think. & Bud. \\
    \midrule
    \multicolumn{7}{l}{\emph{Qwen3.5-2B}} \\
    1 (base) & 80.5 & 31.3 & 14.0 & 85.0 & 28.0 & 12.0 \\
    0.5 & 30.0 & 60.4 ($+93.1$) & 66.0 & 39.8 & 53.7 ($+91.6$) & 58.0 \\
    0.75 & 69.0 & 40.7 ($+30.2$) & 28.0 & 72.4 & 35.3 ($+25.8$) & 25.2 \\
    1.25 & \cellcolor{bestcell}85.5 & 26.4 ($-15.5$) & 9.0 & 85.8 & 25.1 ($-10.6$) & 10.0 \\
    1.5 & 82.0 & 26.8 ($-14.4$) & 9.0 & \cellcolor{bestcell}88.0 & 23.0 ($-18.1$) & 5.6 \\
    1.75 & 84.5 & \cellcolor{bestcell}23.7 ($-24.4$) & 5.0 & 87.0 & 21.9 ($-21.9$) & 5.8 \\
    2 & 80.0 & 24.3 ($-22.3$) & 9.0 & 84.2 & \cellcolor{bestcell}21.3 ($-24.1$) & 6.2 \\
    \midrule
    \multicolumn{7}{l}{\emph{Qwen3.5-4B}} \\
    1 (base) & 98.0 & 10.8 & 0.5 & \cellcolor{bestcell}99.0 & 9.5 & 0.4 \\
    0.5 & 93.5 & 18.7 ($+72.9$) & 4.0 & 94.2 & 17.5 ($+82.9$) & 4.0 \\
    0.75 & 98.0 & 12.8 ($+18.3$) & 1.0 & 98.8 & 11.5 ($+20.8$) & 0.4 \\
    1.25 & 98.0 & 10.0 ($-7.7$) & 0.5 & 98.6 & 8.4 ($-12.1$) & 0.0 \\
    1.5 & \cellcolor{bestcell}98.5 & 8.6 ($-20.8$) & 0.0 & 98.8 & 7.8 ($-18.6$) & 0.0 \\
    1.75 & 97.5 & 8.5 ($-21.7$) & 0.0 & 98.4 & 7.8 ($-18.4$) & 0.0 \\
    2 & 97.0 & \cellcolor{bestcell}8.4 ($-22.4$) & 0.5 & 97.6 & \cellcolor{bestcell}7.7 ($-19.6$) & 0.0 \\
    \midrule
    \multicolumn{7}{l}{\emph{Qwen3.5-9B}} \\
    1 (base) & 98.0 & 9.4 & 0.5 & 99.0 & 8.3 & 0.2 \\
    0.5 & 97.0 & 16.4 ($+73.4$) & 2.5 & 96.8 & 15.2 ($+82.8$) & 2.4 \\
    0.75 & 98.5 & 11.2 ($+18.7$) & 0.5 & \cellcolor{bestcell}99.4 & 9.8 ($+17.9$) & 0.2 \\
    1.25 & \cellcolor{bestcell}99.5 & 8.1 ($-14.4$) & 0.0 & 99.0 & 7.6 ($-8.0$) & 0.2 \\
    1.5 & 98.5 & 7.7 ($-18.2$) & 0.0 & 99.2 & 7.0 ($-15.9$) & 0.0 \\
    1.75 & 98.5 & \cellcolor{bestcell}7.4 ($-22.0$) & 0.0 & 99.2 & 6.8 ($-17.8$) & 0.0 \\
    2 & 97.5 & 7.5 ($-21.1$) & 0.5 & 98.8 & \cellcolor{bestcell}6.6 ($-21.0$) & 0.0 \\
    \bottomrule
  \end{tabular}%
  }
  \caption{Results with a budget of 81,920 tokens for every value of $\kappa$ from 0.5 to 2. $\kappa<1$ forgets more slowly than the unedited model ($\kappa=1$) and $\kappa>1$ faster. Acc.: accuracy (\%). Think.: mean thinking tokens in thousands, with the change from the base model in parentheses (\%). Bud.: share of responses that reach the budget (\%), which count as wrong. Shaded: the highest accuracy and the shortest mean thinking for each model and benchmark (ties shaded together); shading marks the best observed value, not a selected $\kappa$.}
  \label{tab:main82k}
\end{table}

\section{Smaller Token Budgets}
\label{app:budgets}

Because unfinished responses count as wrong, any fixed budget mixes
reasoning length with completion. We therefore reran every model, benchmark
and value of $\kappa$ under budgets of 1,024, 2,048, 4,096, 8,192 and 16,384
tokens. At each budget every question is answered anew, with at most that
many generated tokens, thinking and answer together, and with the settings
and per-question seeds of \S\ref{sec:setup}; one engine per model, setting and shard runs every budget, with a maximum model length of 18,432. A response still thinking at the
budget is unfinished and counts as wrong, and intervals are paired bootstrap
intervals, as in \S\ref{sec:setup}. Table~\ref{tab:budgets} compares
$\kappa=2$ with the base model, and Tables~\ref{tab:budgets-math},
\ref{tab:budgets-gsm8k} and~\ref{tab:budgets-aime} give every value of
$\kappa$.

\begin{table}
  \centering\footnotesize\setlength{\tabcolsep}{3pt}\newcommand{\cen}[1]{\textcolor{gray}{#1}}
  \resizebox{\linewidth}{!}{%
  \begin{tabular}{@{}llrrrrr@{}}
    \toprule
    & & \multicolumn{5}{c}{Token budget} \\
    \cmidrule(lr){3-7}
    Model & Bench. & 2k & 4k & 8k & 16k & 34k \\
    \midrule
    \multicolumn{7}{l}{\emph{$\Delta$Thinking (\%)}} \\
    2B & MATH-Dev & \cen{$+0.8^{*}$} & \cen{$+0.4$} & \cen{$+0.9$} & \cen{$-1.6$} & \cellcolor{bestcell}$-13.7^{*}$ \\
     & GSM8K & \cen{$+1.3^{*}$} & \cen{$-0.5$} & \cen{$+0.3$} & \cellcolor{bestcell}$-7.6^{*}$ & \cellcolor{bestcell}$-18.6^{*}$ \\
     & AIME & \cen{$0.0$} & \cen{$0.0$} & \cen{$0.0$} & \cen{$-0.7$} & \cen{$-3.4$} \\
    4B & MATH-Dev & \cen{$-0.5$} & \cen{$-1.6$} & \cellcolor{bestcell}$-7.5^{*}$ & \cellcolor{bestcell}$-13.0^{*}$ & \cellcolor{bestcell}$-12.2^{*}$ \\
     & GSM8K & \cen{$+2.2^{*}$} & $-2.0$ & \cellcolor{bestcell}$-4.7^{*}$ & \cellcolor{bestcell}$-7.3^{*}$ & \cellcolor{bestcell}$-10.4^{*}$ \\
     & AIME & \cen{$0.0$} & \cen{$-0.4$} & \cen{$-0.9$} & \cen{$-3.2$} & \cellcolor{bestcell}$-17.0^{*}$ \\
    9B & MATH-Dev & \cen{$-1.2$} & \cellcolor{bestcell}\cen{$-3.1^{*}$} & \cellcolor{bestcell}$-7.2^{*}$ & \cellcolor{bestcell}$-11.1^{*}$ & \cellcolor{bestcell}$-22.6^{*}$ \\
     & GSM8K & \cen{$+1.4$} & \cellcolor{bestcell}$-5.0^{*}$ & \cellcolor{bestcell}$-7.4^{*}$ & \cellcolor{bestcell}$-12.1^{*}$ & \cellcolor{bestcell}$-10.4^{*}$ \\
     & AIME & \cen{$0.0$} & \cen{$-0.3$} & \cen{$-0.8$} & \cellcolor{bestcell}\cen{$-5.7^{*}$} & \cellcolor{bestcell}$-12.9^{*}$ \\
    \midrule
    \multicolumn{7}{l}{\emph{$\Delta$Accuracy (pp)}} \\
    2B & MATH-Dev & \cen{$-1.5$} & \cen{$+0.5$} & \cen{$+0.5$} & \cellcolor{bestcell}\cen{$+6.0^{*}$} & \cellcolor{bestcell}$+9.5^{*}$ \\
     & GSM8K & \cen{$-2.1^{*}$} & \cen{$+1.9$} & \cen{$+0.4$} & \cellcolor{bestcell}$+9.2^{*}$ & \cellcolor{bestcell}$+2.7^{*}$ \\
     & AIME & \cen{$0.0$} & \cen{$0.0$} & \cen{$0.0$} & \cen{$+5.0$} & \cen{$+5.0$} \\
    4B & MATH-Dev & \cen{$+2.5$} & \cellcolor{bestcell}\cen{$+6.5^{*}$} & \cellcolor{bestcell}$+10.0^{*}$ & \cellcolor{bestcell}$+12.5^{*}$ & $+0.5$ \\
     & GSM8K & \cen{$-3.9^{*}$} & \cellcolor{bestcell}$+7.5^{*}$ & $+1.2$ & $0.0$ & $-0.4$ \\
     & AIME & \cen{$0.0$} & \cen{$+1.7$} & \cen{$0.0$} & \cen{$+6.7$} & \cellcolor{bestcell}$+20.0^{*}$ \\
    9B & MATH-Dev & \cen{$+3.5$} & \cen{$+4.0$} & \cellcolor{bestcell}$+9.0^{*}$ & \cellcolor{bestcell}$+9.5^{*}$ & $+1.0$ \\
     & GSM8K & \cen{$-3.9^{*}$} & \cellcolor{bestcell}$+7.7^{*}$ & $+1.0$ & $+0.2$ & $-0.6$ \\
     & AIME & \cen{$0.0$} & \cen{$0.0$} & \cen{$+1.7$} & \cellcolor{bestcell}\cen{$+16.7^{*}$} & $+1.7$ \\
    \bottomrule
  \end{tabular}%
  }
  \caption{The edit at $\kappa=2$ against the base model under budgets of 2,048 to 16,384 tokens (2k--16k) and the main budget of 34,000 (34k, Table~\ref{tab:k2}), for Qwen3.5-2B/4B/9B. $\Delta$Thinking: change in mean thinking tokens (\%). $\Delta$Accuracy: change in accuracy (pp). $^{*}$: the 95\% paired bootstrap interval excludes zero. \textcolor{gray}{Grey}: more than half of the base responses are unfinished at that budget. Shaded: the change favours the edit and is marked $^{*}$.}
  \label{tab:budgets}
\end{table}

\paragraph{The saving needs room to finish.}
Mean thinking at a budget averages lengths cut off at that budget, so when
most traces in both settings are cut off, the two means converge whatever the
edit does. At 1,024 tokens, 78--100\% of base responses are unfinished, and
no value of $\kappa$ changes mean thinking by more than 2.4\%. As the budget
grows, the saving at $\kappa=2$ appears and widens (Table~\ref{tab:budgets}).
On MATH-Dev, for the 4B and 9B models, it grows from 1.6\% and 3.1\% at
4,096 tokens to 13.0\% and 11.1\% at 16,384, and is 12.2\% and 22.6\% at
34,000; the
2B model, which leaves 62.5\% of its MATH-Dev responses unfinished even at
16,384 tokens, saves clearly only at 34,000 (13.7\%). On AIME the saving
stays below 1\% up to 8,192 tokens. On GSM8K, which the 4B and 9B models
almost always finish by 16,384 tokens, it is of similar size at 16,384 and
34,000 tokens (7.3\% and 10.4\% for the 4B model, 12.1\% and 10.4\% for the
9B model). In the 11 settings below 34,000 tokens where at most half of the
base responses are unfinished, $\kappa=2$ shortened thinking in all 11
(interval below zero in 10) and $\kappa=0.5$ lengthened it in all 11 (every
interval above zero). The four intervals above zero for $\kappa=2$, each at
most 2.2\%, all lie where most base responses are unfinished: on GSM8K at
1,024 tokens (2B) and 2,048 tokens (2B, 4B) and on MATH-Dev at 2,048 tokens
(2B).

\paragraph{Under a cap, accuracy follows completion.}
In the same 11 settings, $\kappa=2$ raised accuracy in 10 and left it
unchanged in one (4B, GSM8K, 16,384 tokens), with the interval above zero in
7, and $\kappa=0.5$ lowered it in all 11 (every interval below zero). Across
all budgets below 34,000, the largest gain at $\kappa=2$ is on AIME at 16,384 tokens for the 9B model, $+16.7$~pp, where
the share of unfinished responses fell by 16.7~pp (71.7\% to 55.0\%); on
MATH-Dev at 16,384 tokens the 4B model gained 12.5~pp as its unfinished
share fell from 17.0\% to 9.0\%. Across all 450 comparisons (three models,
three benchmarks, five budgets and ten values of $\kappa$), accuracy changes
by about as much as the share of finished responses (slope 0.90~pp per pp,
Pearson $r=0.96$; $r=0.94$ for $\kappa>1$ alone). The four detected losses at
$\kappa=2$ fit this pattern: all are on GSM8K at 1,024 or 2,048 tokens, where
most base responses are unfinished, and in each the edit left more responses
unfinished (for the 4B model at 2,048 tokens, 56.4\% to 59.4\%, with a loss
of 3.9~pp). Together with the transitions on MATH-500 (Appendix~\ref{app:heldout}),
this indicates that under a cap the edit raises accuracy mainly by letting
more traces finish, and that its saving in thinking grows with the room the
budget leaves.

\paragraph{Every value of $\kappa$.}
Slower forgetting lowers accuracy even at 1,024 tokens on GSM8K ($\kappa=0.5$:
$-2.0$, $-1.7$ and $-5.0$~pp for the 2B, 4B and 9B models), because more
responses fail to finish, and it costs more wherever responses can finish:
at 16,384 tokens, $\kappa=0.5$ lowered accuracy by 20.0 and 24.4~pp for the
2B model on MATH-Dev and GSM8K and by 17.0 and 20.0~pp for the 4B and 9B
models on MATH-Dev (Tables~\ref{tab:budgets-math}
and~\ref{tab:budgets-gsm8k}). On AIME, the unedited 2B model finishes no
response below 34,000 tokens, and only one value of $\kappa$ separates from
it there ($\kappa=1.5$ at 16,384 tokens, $+6.7$~pp;
Table~\ref{tab:budgets-aime}).

\begin{table*}
  \centering\footnotesize\setlength{\tabcolsep}{1.9pt}
  \begin{tabular}{@{}lrrrrrrrrrrrrrrr@{}}
    \toprule
    & \multicolumn{3}{c}{1,024} & \multicolumn{3}{c}{2,048} & \multicolumn{3}{c}{4,096} & \multicolumn{3}{c}{8,192} & \multicolumn{3}{c}{16,384} \\
    \cmidrule(lr){2-4}\cmidrule(lr){5-7}\cmidrule(lr){8-10}\cmidrule(lr){11-13}\cmidrule(lr){14-16}
    $\kappa$ & Acc. & Th. & Unf. & Acc. & Th. & Unf. & Acc. & Th. & Unf. & Acc. & Th. & Unf. & Acc. & Th. & Unf. \\
    \midrule
    \multicolumn{16}{l}{\emph{Qwen3.5-2B}} \\
    Base & 0.0 & 1,023 & 99.5 & 2.5 & 2,024 & 97.0 & 6.5 & 3,964 & 92.5 & 17.0 & 7,449 & 81.5 & 34.5 & 13,071 & 62.5 \\
    0.5 & $0.0$ & $+0.1$ & 100.0 & $-1.0$ & $+0.4$ & 98.0 & $-4.5^{*}$ & $+2.2^{*}$ & 96.0 & $-8.5^{*}$ & $+5.3^{*}$ & 91.0 & $-20.0^{*}$ & $+14.8^{*}$ & 84.0 \\
    0.75 & $0.0$ & $+0.1$ & 100.0 & $-1.5$ & $+0.6$ & 98.5 & $+0.5$ & $+0.2$ & 92.0 & $-4.0$ & $+2.2$ & 85.5 & $-6.0^{*}$ & $+6.7^{*}$ & 70.0 \\
    1.25 & $0.0$ & $+0.1$ & 99.5 & $-1.5$ & $+0.5$ & 98.0 & $+1.5$ & $-0.8$ & 90.5 & $+2.5$ & $-1.2$ & 79.0 & $+3.5$ & $+0.7$ & 59.0 \\
    1.5 & $0.0$ & $0.0$ & 99.0 & $-1.5$ & $+0.5$ & 97.5 & \cellcolor{bestcell}$+3.5$ & \cellcolor{bestcell}$-1.0$ & 88.0 & \cellcolor{bestcell}$+4.0$ & $-1.2$ & 78.0 & $+4.5$ & $-1.1$ & 57.5 \\
    1.75 & $0.0$ & $+0.1$ & 100.0 & $-1.5$ & $+0.7$ & 98.5 & $+3.0$ & $-0.7$ & 88.5 & $+2.0$ & \cellcolor{bestcell}$-1.3$ & 76.5 & \cellcolor{bestcell}$+10.0^{*}$ & \cellcolor{bestcell}$-2.7$ & 52.0 \\
    2 & $0.0$ & $+0.1$ & 100.0 & $-1.5$ & $+0.8^{*}$ & 98.0 & $+0.5$ & $+0.4$ & 91.0 & $+0.5$ & $+0.9$ & 77.5 & $+6.0^{*}$ & $-1.6$ & 56.0 \\
    2.25 & $0.0$ & $+0.1$ & 99.0 & \cellcolor{bestcell}$+0.5$ & \cellcolor{bestcell}$-0.1$ & 96.0 & $+1.0$ & $0.0$ & 90.5 & $0.0$ & $+0.7$ & 81.0 & $+5.5$ & $+0.6$ & 55.5 \\
    2.5 & $0.0$ & $+0.1$ & 100.0 & $-0.5$ & $+0.6$ & 97.0 & $0.0$ & $+0.1$ & 92.5 & $+0.5$ & $-0.2$ & 80.0 & $+5.0$ & $+0.9$ & 53.5 \\
    2.75 & $0.0$ & $+0.1$ & 100.0 & $-2.5^{*}$ & $+1.2^{*}$ & 100.0 & $0.0$ & $-0.3$ & 91.5 & $-4.5$ & $+2.2$ & 83.5 & $+5.5$ & $-0.2$ & 51.0 \\
    3 & $0.0$ & $+0.1$ & 99.5 & $-1.0$ & $+0.8^{*}$ & 98.0 & $-2.0$ & $+1.2$ & 94.5 & $-1.0$ & $+2.3^{*}$ & 81.5 & $-0.5$ & $+4.0^{*}$ & 61.0 \\
    \midrule
    \multicolumn{16}{l}{\emph{Qwen3.5-4B}} \\
    Base & 0.0 & 1,023 & 99.0 & 4.0 & 1,997 & 92.0 & 21.0 & 3,668 & 71.5 & 52.0 & 5,986 & 43.0 & 76.5 & 8,411 & 17.0 \\
    0.5 & $0.0$ & $-0.1$ & 98.5 & $-1.5$ & $+0.9^{*}$ & 95.0 & $-5.5^{*}$ & $+3.7^{*}$ & 82.0 & $-15.0^{*}$ & $+12.0^{*}$ & 58.5 & $-17.0^{*}$ & $+24.4^{*}$ & 37.0 \\
    0.75 & $0.0$ & $-0.1$ & 98.5 & $-1.5$ & $+0.9^{*}$ & 94.0 & $-1.0$ & $+1.4$ & 77.0 & $-3.0$ & $+1.8$ & 45.5 & $-7.0^{*}$ & $+9.3^{*}$ & 27.5 \\
    1.25 & $0.0$ & \cellcolor{bestcell}$-0.2^{*}$ & 98.0 & $+1.5$ & $-0.2$ & 90.5 & $+5.5^{*}$ & $-1.6$ & 66.0 & $+3.5$ & $-2.7$ & 40.0 & $+6.5^{*}$ & $-3.4$ & 14.0 \\
    1.5 & \cellcolor{bestcell}$+0.5$ & $-0.1$ & 98.5 & $+0.5$ & $-0.1$ & 90.5 & $+5.5^{*}$ & $-2.7^{*}$ & 66.0 & $+2.5$ & $-3.4^{*}$ & 39.0 & $+7.0^{*}$ & $-9.0^{*}$ & 12.0 \\
    1.75 & $0.0$ & $0.0$ & 98.5 & $+0.5$ & $+0.1$ & 90.0 & $+5.5^{*}$ & $-2.2$ & 66.5 & $+5.5^{*}$ & $-5.7^{*}$ & 37.5 & $+10.5^{*}$ & $-12.3^{*}$ & 9.5 \\
    2 & $0.0$ & \cellcolor{bestcell}$-0.2$ & 97.0 & \cellcolor{bestcell}$+2.5$ & $-0.5$ & 90.5 & $+6.5^{*}$ & $-1.6$ & 65.0 & \cellcolor{bestcell}$+10.0^{*}$ & $-7.5^{*}$ & 33.5 & \cellcolor{bestcell}$+12.5^{*}$ & $-13.0^{*}$ & 9.0 \\
    2.25 & $0.0$ & $0.0$ & 99.0 & \cellcolor{bestcell}$+2.5$ & \cellcolor{bestcell}$-0.9$ & 87.0 & $+6.5^{*}$ & $-2.2^{*}$ & 67.0 & \cellcolor{bestcell}$+10.0^{*}$ & $-8.6^{*}$ & 34.5 & $+9.5^{*}$ & $-13.0^{*}$ & 11.5 \\
    2.5 & \cellcolor{bestcell}$+0.5$ & $-0.1$ & 99.0 & $+1.0$ & $+0.3$ & 90.0 & \cellcolor{bestcell}$+8.0^{*}$ & $-2.0$ & 64.0 & \cellcolor{bestcell}$+10.0^{*}$ & \cellcolor{bestcell}$-9.0^{*}$ & 31.0 & $+11.0^{*}$ & \cellcolor{bestcell}$-15.1^{*}$ & 10.0 \\
    2.75 & $0.0$ & $-0.1$ & 98.0 & $-1.5$ & $+1.1$ & 92.5 & $+6.0^{*}$ & $-1.2$ & 67.5 & $+6.0^{*}$ & $-5.8^{*}$ & 37.5 & $+10.0^{*}$ & $-13.0^{*}$ & 11.5 \\
    3 & \cellcolor{bestcell}$+0.5$ & $0.0$ & 99.5 & $+2.0$ & $-0.3$ & 90.5 & \cellcolor{bestcell}$+8.0^{*}$ & \cellcolor{bestcell}$-3.5^{*}$ & 65.0 & $+8.5^{*}$ & $-7.1^{*}$ & 35.5 & $+10.0^{*}$ & $-11.5^{*}$ & 11.0 \\
    \midrule
    \multicolumn{16}{l}{\emph{Qwen3.5-9B}} \\
    Base & 0.0 & 1,023 & 99.5 & 3.5 & 2,008 & 90.0 & 25.0 & 3,624 & 68.5 & 57.0 & 5,752 & 38.5 & 82.0 & 7,673 & 15.0 \\
    0.5 & \cellcolor{bestcell}$+0.5$ & $0.0$ & 99.0 & $+1.0$ & $+0.1$ & 92.0 & $-8.0^{*}$ & $+4.3^{*}$ & 77.5 & $-14.5^{*}$ & $+11.9^{*}$ & 53.0 & $-20.0^{*}$ & $+30.1^{*}$ & 34.0 \\
    0.75 & \cellcolor{bestcell}$+0.5$ & $-0.1$ & 98.5 & $+2.5$ & \cellcolor{bestcell}$-1.3^{*}$ & 88.5 & $-6.5^{*}$ & $+2.6$ & 74.5 & $-6.0^{*}$ & $+3.2$ & 40.5 & $-5.5^{*}$ & $+10.3^{*}$ & 20.5 \\
    1.25 & $0.0$ & $0.0$ & 98.5 & $+0.5$ & $-0.5$ & 90.5 & $-2.0$ & $-1.0$ & 66.5 & $+0.5$ & $-3.7^{*}$ & 36.5 & $+4.5^{*}$ & $-5.3^{*}$ & 11.0 \\
    1.5 & $0.0$ & $+0.1$ & 99.0 & $+2.0$ & $-0.8$ & 91.0 & $+1.0$ & $+0.5$ & 64.5 & $+6.0^{*}$ & $-5.7^{*}$ & 30.0 & $+7.5^{*}$ & $-9.1^{*}$ & 9.0 \\
    1.75 & \cellcolor{bestcell}$+0.5$ & \cellcolor{bestcell}$-0.5$ & 98.0 & $+1.5$ & $-0.9$ & 90.5 & $+2.0$ & $-2.2$ & 67.5 & $+6.5^{*}$ & $-6.6^{*}$ & 32.5 & $+7.5^{*}$ & $-13.1^{*}$ & 5.0 \\
    2 & \cellcolor{bestcell}$+0.5$ & $-0.3$ & 97.5 & \cellcolor{bestcell}$+3.5$ & $-1.2$ & 89.0 & $+4.0$ & $-3.1^{*}$ & 60.0 & $+9.0^{*}$ & $-7.2^{*}$ & 28.0 & $+9.5^{*}$ & $-11.1^{*}$ & 8.0 \\
    2.25 & $0.0$ & $-0.2$ & 98.5 & \cellcolor{bestcell}$+3.5$ & $-1.2$ & 88.5 & $+2.5$ & $-2.0$ & 62.0 & $+7.5^{*}$ & $-5.2^{*}$ & 31.0 & $+9.5^{*}$ & $-14.2^{*}$ & 8.0 \\
    2.5 & \cellcolor{bestcell}$+0.5$ & $-0.2$ & 98.5 & $+2.5$ & $-0.7$ & 89.5 & $+2.5$ & $-1.7$ & 66.5 & $+8.5^{*}$ & $-7.9^{*}$ & 26.0 & \cellcolor{bestcell}$+10.0^{*}$ & $-15.8^{*}$ & 6.5 \\
    2.75 & \cellcolor{bestcell}$+0.5$ & $-0.2$ & 98.5 & $0.0$ & $0.0$ & 90.5 & \cellcolor{bestcell}$+6.0^{*}$ & \cellcolor{bestcell}$-3.4^{*}$ & 62.0 & $+11.5^{*}$ & \cellcolor{bestcell}$-9.2^{*}$ & 27.0 & $+9.5^{*}$ & $-15.0^{*}$ & 8.0 \\
    3 & $0.0$ & $0.0$ & 99.0 & $+1.0$ & $-0.5$ & 91.0 & $+3.5$ & $-1.4$ & 65.5 & \cellcolor{bestcell}$+12.0^{*}$ & $-8.2^{*}$ & 26.0 & $+8.5^{*}$ & \cellcolor{bestcell}$-17.3^{*}$ & 7.0 \\
    \bottomrule
  \end{tabular}
  \caption{MATH-Dev (levels 3--5, 200 problems) under token budgets of 1,024 to 16,384 tokens, for every value of $\kappa$. Base rows: accuracy (Acc., \%), mean thinking tokens (Th.) and the share of responses still thinking at the budget (Unf., \%), which count as wrong. Other rows: change in accuracy (pp) and in mean thinking tokens (\%) against the base model at the same budget, and the share unfinished (\%). $^{*}$: the 95\% paired bootstrap interval excludes zero. Shaded: for each model and budget, the edit with the largest accuracy gain and the edit with the largest reduction in thinking, when it improves on the base model (ties shaded together).}
  \label{tab:budgets-math}
\end{table*}
\begin{table*}
  \centering\footnotesize\setlength{\tabcolsep}{1.9pt}
  \begin{tabular}{@{}lrrrrrrrrrrrrrrr@{}}
    \toprule
    & \multicolumn{3}{c}{1,024} & \multicolumn{3}{c}{2,048} & \multicolumn{3}{c}{4,096} & \multicolumn{3}{c}{8,192} & \multicolumn{3}{c}{16,384} \\
    \cmidrule(lr){2-4}\cmidrule(lr){5-7}\cmidrule(lr){8-10}\cmidrule(lr){11-13}\cmidrule(lr){14-16}
    $\kappa$ & Acc. & Th. & Unf. & Acc. & Th. & Unf. & Acc. & Th. & Unf. & Acc. & Th. & Unf. & Acc. & Th. & Unf. \\
    \midrule
    \multicolumn{16}{l}{\emph{Qwen3.5-2B}} \\
    Base & 3.3 & 1,008 & 94.2 & 12.8 & 1,918 & 84.7 & 25.5 & 3,539 & 71.1 & 47.1 & 5,920 & 51.2 & 65.1 & 9,337 & 32.6 \\
    0.5 & $-2.0^{*}$ & $+1.0^{*}$ & 97.6 & $-7.1^{*}$ & $+3.8^{*}$ & 92.7 & $-12.4^{*}$ & $+8.0^{*}$ & 85.7 & $-19.6^{*}$ & $+18.3^{*}$ & 71.4 & $-24.4^{*}$ & $+32.1^{*}$ & 57.3 \\
    0.75 & $-1.1$ & $+0.4$ & 94.6 & $-1.7$ & $+1.4^{*}$ & 87.6 & $-3.5^{*}$ & $+1.8^{*}$ & 75.9 & $-7.1^{*}$ & $+7.2^{*}$ & 58.0 & $-7.1^{*}$ & $+9.3^{*}$ & 39.8 \\
    1.25 & $-0.5$ & $+0.1$ & 93.8 & \cellcolor{bestcell}$+2.0$ & \cellcolor{bestcell}$-0.6$ & 82.8 & \cellcolor{bestcell}$+2.7^{*}$ & \cellcolor{bestcell}$-1.5$ & 68.7 & $+2.7^{*}$ & \cellcolor{bestcell}$-1.6$ & 48.3 & $+4.5^{*}$ & $-4.6^{*}$ & 27.7 \\
    1.5 & $-0.4$ & $+0.1$ & 94.3 & $-0.5$ & $+0.3$ & 85.3 & $+2.2$ & $-1.0$ & 69.5 & \cellcolor{bestcell}$+2.9^{*}$ & $-1.0$ & 47.2 & $+7.0^{*}$ & $-6.9^{*}$ & 24.5 \\
    1.75 & $-0.8$ & $+0.1$ & 94.0 & $-1.3$ & $+0.8$ & 84.9 & $+2.5^{*}$ & $-0.2$ & 69.2 & $+2.3$ & $-0.4$ & 48.1 & $+7.7^{*}$ & \cellcolor{bestcell}$-8.4^{*}$ & 23.6 \\
    2 & $-1.4^{*}$ & $+0.5^{*}$ & 95.3 & $-2.1^{*}$ & $+1.3^{*}$ & 86.4 & $+1.9$ & $-0.5$ & 71.0 & $+0.4$ & $+0.3$ & 49.7 & \cellcolor{bestcell}$+9.2^{*}$ & $-7.6^{*}$ & 21.1 \\
    2.25 & $-2.2^{*}$ & $+0.9^{*}$ & 96.9 & $-4.2^{*}$ & $+2.7^{*}$ & 87.9 & $-1.1$ & $+2.0^{*}$ & 73.2 & $+0.2$ & $+3.6^{*}$ & 50.0 & $+8.7^{*}$ & $-6.2^{*}$ & 21.5 \\
    2.5 & $-1.9^{*}$ & $+0.9^{*}$ & 97.3 & $-4.1^{*}$ & $+2.5^{*}$ & 89.0 & $-3.2^{*}$ & $+3.1^{*}$ & 75.2 & $-0.8$ & $+4.2^{*}$ & 49.7 & $+7.9^{*}$ & $-5.0^{*}$ & 20.6 \\
    2.75 & $-2.5^{*}$ & $+1.1^{*}$ & 98.0 & $-4.5^{*}$ & $+2.9^{*}$ & 89.3 & $-6.2^{*}$ & $+4.4^{*}$ & 78.7 & $-2.7^{*}$ & $+7.3^{*}$ & 51.9 & $+8.9^{*}$ & $-3.7^{*}$ & 18.1 \\
    3 & $-2.5^{*}$ & $+1.1^{*}$ & 98.0 & $-5.5^{*}$ & $+3.2^{*}$ & 90.4 & $-8.4^{*}$ & $+5.7^{*}$ & 81.0 & $-1.4$ & $+8.4^{*}$ & 49.4 & $+8.7^{*}$ & $-3.4^{*}$ & 16.1 \\
    \midrule
    \multicolumn{16}{l}{\emph{Qwen3.5-4B}} \\
    Base & 8.0 & 985 & 81.4 & 37.1 & 1,671 & 56.4 & 61.3 & 2,609 & 33.9 & 87.5 & 3,312 & 10.0 & 93.9 & 3,798 & 3.0 \\
    0.5 & $-1.7^{*}$ & $+1.5^{*}$ & 88.3 & $-7.0^{*}$ & $+5.8^{*}$ & 66.3 & $-11.1^{*}$ & $+12.2^{*}$ & 46.8 & $-9.6^{*}$ & $+27.4^{*}$ & 19.8 & $-4.4^{*}$ & $+35.6^{*}$ & 7.7 \\
    0.75 & $-1.1$ & $+0.7^{*}$ & 83.8 & $-2.4$ & $+2.5^{*}$ & 60.4 & $-3.6^{*}$ & $+3.8^{*}$ & 38.4 & $-3.0^{*}$ & $+11.1^{*}$ & 13.3 & $-1.4^{*}$ & $+10.9^{*}$ & 4.5 \\
    1.25 & \cellcolor{bestcell}$+0.2$ & $-0.2$ & 80.9 & \cellcolor{bestcell}$+1.1$ & \cellcolor{bestcell}$-0.9$ & 56.2 & $+3.8^{*}$ & \cellcolor{bestcell}$-2.8^{*}$ & 30.4 & \cellcolor{bestcell}$+1.7^{*}$ & $-3.8^{*}$ & 8.0 & $+0.4$ & $-5.3^{*}$ & 2.1 \\
    1.5 & $-0.3$ & \cellcolor{bestcell}$-0.4$ & 80.1 & $-0.1$ & $0.0$ & 57.4 & $+5.2^{*}$ & \cellcolor{bestcell}$-2.8^{*}$ & 28.6 & $+0.3$ & $-3.9^{*}$ & 9.1 & $+0.1$ & $-6.7^{*}$ & 2.8 \\
    1.75 & $-0.1$ & $0.0$ & 83.0 & $-3.6^{*}$ & $+1.8^{*}$ & 60.0 & $+5.6^{*}$ & $-2.5$ & 27.6 & $+1.0$ & $-4.4^{*}$ & 8.0 & \cellcolor{bestcell}$+0.5$ & \cellcolor{bestcell}$-9.9^{*}$ & 1.8 \\
    2 & $-0.1$ & $-0.1$ & 83.0 & $-3.9^{*}$ & $+2.2^{*}$ & 59.4 & \cellcolor{bestcell}$+7.5^{*}$ & $-2.0$ & 25.9 & $+1.2$ & \cellcolor{bestcell}$-4.7^{*}$ & 8.3 & $0.0$ & $-7.3^{*}$ & 2.3 \\
    2.25 & $-2.1^{*}$ & $+0.6$ & 85.4 & $-5.9^{*}$ & $+3.8^{*}$ & 60.3 & \cellcolor{bestcell}$+7.5^{*}$ & $-0.3$ & 26.4 & $+1.1$ & $-3.0$ & 8.5 & $+0.1$ & $-5.3^{*}$ & 2.1 \\
    2.5 & $-1.8^{*}$ & $+0.7^{*}$ & 85.6 & $-8.2^{*}$ & $+5.3^{*}$ & 63.3 & $+6.0^{*}$ & $+0.7$ & 27.4 & $+1.1$ & $-1.7$ & 8.4 & $-0.6$ & $-2.8$ & 3.0 \\
    2.75 & $-3.6^{*}$ & $+1.3^{*}$ & 88.7 & $-8.9^{*}$ & $+5.9^{*}$ & 63.7 & $+4.9^{*}$ & $+1.9$ & 28.8 & $+1.4$ & $+1.2$ & 7.6 & $+0.2$ & $-5.0^{*}$ & 1.7 \\
    3 & $-3.3^{*}$ & $+1.4^{*}$ & 89.2 & $-8.0^{*}$ & $+6.5^{*}$ & 62.6 & $+7.1^{*}$ & $+1.3$ & 27.7 & $-0.7$ & $+2.4$ & 9.6 & $-0.3$ & $-0.1$ & 2.4 \\
    \midrule
    \multicolumn{16}{l}{\emph{Qwen3.5-9B}} \\
    Base & 10.5 & 974 & 78.0 & 43.8 & 1,599 & 50.1 & 69.2 & 2,374 & 26.0 & 91.1 & 2,865 & 6.2 & 95.4 & 3,232 & 1.7 \\
    0.5 & $-5.0^{*}$ & $+2.4^{*}$ & 88.3 & $-12.5^{*}$ & $+9.5^{*}$ & 63.9 & $-16.6^{*}$ & $+18.6^{*}$ & 43.6 & $-9.9^{*}$ & $+39.1^{*}$ & 16.2 & $-3.0^{*}$ & $+44.1^{*}$ & 5.3 \\
    0.75 & $-3.6^{*}$ & $+1.5^{*}$ & 82.4 & $-4.6^{*}$ & $+3.8^{*}$ & 54.6 & $-4.5^{*}$ & $+6.3^{*}$ & 32.1 & $-2.4^{*}$ & $+15.8^{*}$ & 8.6 & $0.0$ & $+11.2^{*}$ & 2.1 \\
    1.25 & \cellcolor{bestcell}$+1.2$ & \cellcolor{bestcell}$-1.1^{*}$ & 74.7 & \cellcolor{bestcell}$+0.3$ & $-0.7$ & 50.3 & $+4.4^{*}$ & $-4.4^{*}$ & 21.8 & $+1.2$ & $-2.9$ & 4.8 & $+0.5$ & $-8.7^{*}$ & 1.1 \\
    1.5 & $-0.2$ & $-0.5$ & 75.5 & $+0.2$ & \cellcolor{bestcell}$-0.8$ & 49.4 & $+5.6^{*}$ & $-4.0^{*}$ & 20.4 & $+1.4^{*}$ & $-6.5^{*}$ & 4.9 & \cellcolor{bestcell}$+0.7$ & $-11.2^{*}$ & 0.6 \\
    1.75 & $+0.4$ & $-0.6$ & 75.8 & $-2.1$ & $+1.1$ & 52.2 & $+5.9^{*}$ & $-3.7^{*}$ & 20.5 & $+1.7^{*}$ & $-4.3^{*}$ & 4.6 & $+0.4$ & $-10.1^{*}$ & 0.9 \\
    2 & $-0.8$ & $-0.2$ & 77.9 & $-3.9^{*}$ & $+1.4$ & 52.9 & \cellcolor{bestcell}$+7.7^{*}$ & \cellcolor{bestcell}$-5.0^{*}$ & 18.6 & $+1.0$ & $-7.4^{*}$ & 4.9 & $+0.2$ & $-12.1^{*}$ & 0.9 \\
    2.25 & $+0.4$ & $-0.6$ & 78.0 & $-4.9^{*}$ & $+2.5^{*}$ & 52.5 & $+7.1^{*}$ & $-3.2^{*}$ & 18.8 & $+1.4^{*}$ & \cellcolor{bestcell}$-7.9^{*}$ & 3.3 & $+0.1$ & $-11.5^{*}$ & 1.1 \\
    2.5 & $-0.4$ & $+0.1$ & 80.7 & $-3.8^{*}$ & $+2.0^{*}$ & 51.2 & $+7.3^{*}$ & $-1.9$ & 18.3 & \cellcolor{bestcell}$+1.8^{*}$ & $-6.3^{*}$ & 4.0 & $0.0$ & $-11.7^{*}$ & 0.8 \\
    2.75 & $-0.8$ & $-0.2$ & 80.0 & $-6.3^{*}$ & $+4.0^{*}$ & 54.1 & \cellcolor{bestcell}$+7.7^{*}$ & $-2.0$ & 18.4 & $+1.4^{*}$ & $-5.4^{*}$ & 4.7 & $0.0$ & \cellcolor{bestcell}$-12.6^{*}$ & 0.8 \\
    3 & $-2.5^{*}$ & $+0.8^{*}$ & 83.4 & $-5.3^{*}$ & $+3.9^{*}$ & 53.7 & $+7.0^{*}$ & $-2.0$ & 18.8 & $+0.5$ & $-3.6$ & 5.4 & $-0.3$ & $-8.3^{*}$ & 1.1 \\
    \bottomrule
  \end{tabular}
  \caption{GSM8K (1,319 problems) under token budgets of 1,024 to 16,384 tokens. Columns and shading as in Table~\ref{tab:budgets-math}.}
  \label{tab:budgets-gsm8k}
\end{table*}
\begin{table*}
  \centering\footnotesize\setlength{\tabcolsep}{1.9pt}
  \begin{tabular}{@{}lrrrrrrrrrrrrrrr@{}}
    \toprule
    & \multicolumn{3}{c}{1,024} & \multicolumn{3}{c}{2,048} & \multicolumn{3}{c}{4,096} & \multicolumn{3}{c}{8,192} & \multicolumn{3}{c}{16,384} \\
    \cmidrule(lr){2-4}\cmidrule(lr){5-7}\cmidrule(lr){8-10}\cmidrule(lr){11-13}\cmidrule(lr){14-16}
    $\kappa$ & Acc. & Th. & Unf. & Acc. & Th. & Unf. & Acc. & Th. & Unf. & Acc. & Th. & Unf. & Acc. & Th. & Unf. \\
    \midrule
    \multicolumn{16}{l}{\emph{Qwen3.5-2B}} \\
    Base & 0.0 & 1,024 & 100.0 & 0.0 & 2,048 & 100.0 & 0.0 & 4,096 & 100.0 & 0.0 & 8,192 & 100.0 & 0.0 & 16,384 & 100.0 \\
    0.5 & $0.0$ & $0.0$ & 100.0 & $0.0$ & $0.0$ & 100.0 & $0.0$ & $0.0$ & 100.0 & $0.0$ & $0.0$ & 100.0 & $0.0$ & $0.0$ & 100.0 \\
    0.75 & $0.0$ & $0.0$ & 100.0 & $0.0$ & $0.0$ & 100.0 & $0.0$ & $0.0$ & 100.0 & $0.0$ & $0.0$ & 100.0 & $0.0$ & $0.0$ & 100.0 \\
    1.25 & $0.0$ & $0.0$ & 100.0 & $0.0$ & $0.0$ & 100.0 & $0.0$ & $0.0$ & 100.0 & $0.0$ & $0.0$ & 98.3 & $+1.7$ & $-0.3$ & 98.3 \\
    1.5 & $0.0$ & $0.0$ & 100.0 & $0.0$ & $0.0$ & 100.0 & $0.0$ & $0.0$ & 100.0 & $0.0$ & \cellcolor{bestcell}$-0.5$ & 98.3 & \cellcolor{bestcell}$+6.7^{*}$ & \cellcolor{bestcell}$-2.1^{*}$ & 90.0 \\
    1.75 & $0.0$ & $0.0$ & 100.0 & $0.0$ & $0.0$ & 100.0 & $0.0$ & $0.0$ & 100.0 & $0.0$ & $0.0$ & 100.0 & $0.0$ & $-0.1$ & 96.7 \\
    2 & $0.0$ & $0.0$ & 100.0 & $0.0$ & $0.0$ & 100.0 & $0.0$ & $0.0$ & 100.0 & $0.0$ & $0.0$ & 100.0 & $+5.0$ & $-0.7$ & 95.0 \\
    2.25 & $0.0$ & $0.0$ & 100.0 & $0.0$ & $0.0$ & 100.0 & $0.0$ & $0.0$ & 100.0 & $0.0$ & $0.0$ & 100.0 & $+3.3$ & $-0.4$ & 96.7 \\
    2.5 & $0.0$ & $0.0$ & 100.0 & $0.0$ & $0.0$ & 100.0 & $0.0$ & $0.0$ & 100.0 & $0.0$ & $0.0$ & 100.0 & $+5.0$ & $-1.5^{*}$ & 93.3 \\
    2.75 & $0.0$ & $0.0$ & 100.0 & $0.0$ & $0.0$ & 100.0 & $0.0$ & $0.0$ & 100.0 & $0.0$ & $0.0$ & 100.0 & $+3.3$ & $-0.7^{*}$ & 93.3 \\
    3 & $0.0$ & $0.0$ & 100.0 & $0.0$ & $0.0$ & 100.0 & $0.0$ & $0.0$ & 100.0 & $0.0$ & $0.0$ & 100.0 & $0.0$ & $-0.2$ & 95.0 \\
    \midrule
    \multicolumn{16}{l}{\emph{Qwen3.5-4B}} \\
    Base & 0.0 & 1,024 & 100.0 & 0.0 & 2,048 & 100.0 & 0.0 & 4,096 & 100.0 & 5.0 & 8,051 & 91.7 & 23.3 & 14,990 & 75.0 \\
    0.5 & $0.0$ & $0.0$ & 100.0 & $0.0$ & $0.0$ & 100.0 & $0.0$ & $0.0$ & 100.0 & $-3.3$ & $+1.1^{*}$ & 98.3 & $-15.0^{*}$ & $+5.8^{*}$ & 91.7 \\
    0.75 & $0.0$ & $0.0$ & 100.0 & $0.0$ & $0.0$ & 100.0 & $0.0$ & $-0.2$ & 98.3 & $0.0$ & $+0.6$ & 95.0 & $-10.0^{*}$ & $+3.7^{*}$ & 85.0 \\
    1.25 & $0.0$ & $0.0$ & 100.0 & $0.0$ & $0.0$ & 100.0 & $0.0$ & $0.0$ & 98.3 & $-3.3$ & $+1.3$ & 98.3 & $+1.7$ & $+0.2$ & 71.7 \\
    1.5 & $0.0$ & $0.0$ & 100.0 & $0.0$ & $0.0$ & 100.0 & $0.0$ & $0.0$ & 100.0 & $+1.7$ & $0.0$ & 88.3 & $+6.7^{*}$ & $-3.2^{*}$ & 68.3 \\
    1.75 & $0.0$ & $0.0$ & 100.0 & $0.0$ & $0.0$ & 100.0 & $0.0$ & $-0.1$ & 98.3 & \cellcolor{bestcell}$+3.3$ & $-0.7$ & 91.7 & $+3.3$ & $-2.7^{*}$ & 68.3 \\
    2 & $0.0$ & $0.0$ & 100.0 & $0.0$ & $0.0$ & 100.0 & \cellcolor{bestcell}$+1.7$ & \cellcolor{bestcell}$-0.4$ & 96.7 & $0.0$ & $-0.9$ & 90.0 & $+6.7$ & $-3.2$ & 65.0 \\
    2.25 & $0.0$ & $0.0$ & 100.0 & $0.0$ & $0.0$ & 100.0 & $0.0$ & $0.0$ & 100.0 & \cellcolor{bestcell}$+3.3$ & \cellcolor{bestcell}$-1.0$ & 88.3 & $+8.3$ & $-2.5$ & 63.3 \\
    2.5 & $0.0$ & $0.0$ & 100.0 & $0.0$ & $0.0$ & 100.0 & $0.0$ & $0.0$ & 100.0 & $-3.3$ & $+1.0^{*}$ & 96.7 & \cellcolor{bestcell}$+18.3^{*}$ & $-6.3^{*}$ & 56.7 \\
    2.75 & $0.0$ & $0.0$ & 100.0 & $0.0$ & $0.0$ & 100.0 & $0.0$ & $0.0$ & 100.0 & $-1.7$ & $+0.7$ & 95.0 & $+15.0^{*}$ & \cellcolor{bestcell}$-6.7^{*}$ & 58.3 \\
    3 & $0.0$ & $0.0$ & 100.0 & $0.0$ & $0.0$ & 100.0 & $0.0$ & $0.0$ & 100.0 & $+1.7$ & $-0.6$ & 91.7 & $+10.0^{*}$ & $-6.6^{*}$ & 63.3 \\
    \midrule
    \multicolumn{16}{l}{\emph{Qwen3.5-9B}} \\
    Base & 0.0 & 1,024 & 100.0 & 0.0 & 2,048 & 100.0 & 0.0 & 4,096 & 100.0 & 5.0 & 8,022 & 91.7 & 23.3 & 14,623 & 71.7 \\
    0.5 & $0.0$ & $0.0$ & 100.0 & $0.0$ & $0.0$ & 100.0 & $0.0$ & $0.0$ & 100.0 & $-3.3$ & $+1.3$ & 96.7 & $-18.3^{*}$ & $+10.1^{*}$ & 93.3 \\
    0.75 & $0.0$ & $0.0$ & 100.0 & $0.0$ & $0.0$ & 100.0 & $0.0$ & $0.0$ & 98.3 & $-5.0$ & $+1.9^{*}$ & 98.3 & $-6.7^{*}$ & $+5.6^{*}$ & 80.0 \\
    1.25 & $0.0$ & $0.0$ & 100.0 & $0.0$ & $0.0$ & 100.0 & $0.0$ & $0.0$ & 100.0 & $-1.7$ & $+0.3$ & 91.7 & $+6.7$ & $-0.7$ & 66.7 \\
    1.5 & $0.0$ & $0.0$ & 100.0 & $0.0$ & $0.0$ & 100.0 & $0.0$ & $0.0$ & 98.3 & $+3.3$ & $-0.7$ & 88.3 & $+15.0^{*}$ & $-3.9^{*}$ & 60.0 \\
    1.75 & $0.0$ & $0.0$ & 100.0 & $0.0$ & $0.0$ & 100.0 & $0.0$ & $0.0$ & 100.0 & $+1.7$ & $-0.3$ & 88.3 & $+18.3^{*}$ & $-6.0^{*}$ & 55.0 \\
    2 & $0.0$ & $0.0$ & 100.0 & $0.0$ & $0.0$ & 100.0 & $0.0$ & $-0.3$ & 98.3 & $+1.7$ & $-0.8$ & 86.7 & $+16.7^{*}$ & $-5.7^{*}$ & 55.0 \\
    2.25 & $0.0$ & $0.0$ & 100.0 & $0.0$ & $0.0$ & 100.0 & \cellcolor{bestcell}$+1.7$ & \cellcolor{bestcell}$-0.5$ & 98.3 & $+1.7$ & $-0.8$ & 86.7 & \cellcolor{bestcell}$+20.0^{*}$ & $-7.7^{*}$ & 51.7 \\
    2.5 & $0.0$ & $0.0$ & 100.0 & $0.0$ & $0.0$ & 100.0 & $0.0$ & $0.0$ & 100.0 & $+5.0$ & $-1.5^{*}$ & 83.3 & $+18.3^{*}$ & $-9.2^{*}$ & 51.7 \\
    2.75 & $0.0$ & $0.0$ & 100.0 & $0.0$ & $0.0$ & 100.0 & \cellcolor{bestcell}$+1.7$ & $-0.4$ & 96.7 & \cellcolor{bestcell}$+8.3^{*}$ & \cellcolor{bestcell}$-1.8^{*}$ & 83.3 & $+18.3^{*}$ & $-7.7^{*}$ & 53.3 \\
    3 & $0.0$ & $0.0$ & 100.0 & $0.0$ & $0.0$ & 100.0 & $0.0$ & $0.0$ & 98.3 & \cellcolor{bestcell}$+8.3^{*}$ & $-1.7^{*}$ & 85.0 & $+16.7^{*}$ & \cellcolor{bestcell}$-9.4^{*}$ & 58.3 \\
    \bottomrule
  \end{tabular}
  \caption{AIME 2024 and 2025 (60 problems) under token budgets of 1,024 to 16,384 tokens. Columns and shading as in Table~\ref{tab:budgets-math}.}
  \label{tab:budgets-aime}
\end{table*}

\section{Proofs of the Properties of the Edit}
\label{app:proofs}

This appendix proves the properties of the edit stated in
\S\ref{sec:method} and adds results that bound what those properties
imply. The results are elementary, and most restate known facts about gated
linear recurrences, such as the unrolled form of the state
\citep{ali2025hidden,yang2025gdn} and the scale invariance of the RMS norm
\citep{zhang2019rmsnorm}; we collect them, applied to the edit, with the
assumptions under which each one holds. Throughout, we compare one Gated DeltaNet head with and without the
edit $E_\kappa$ while its inputs are held fixed: the clock $\delta_t$, the
write strength $\beta_t$, the key $k_t$, the value $v_t$ and the query $q_t$
are the same in both runs. We drop the head index $h$. As in Qwen3.5, keys
have unit $L_2$ norm and $\beta_t\in[0,1]$. A superscript $(\kappa)$ marks
the edited head, and both heads start from the same state $S_0$.

\paragraph{Notation.}
Write $B_t=I-\beta_t k_t k_t^{\top}$ and $W_t=\beta_t v_t k_t^{\top}$, so
that Eq.~\ref{eq:gdn} reads $S_t=\alpha_t S_{t-1}B_t+W_t$. For $s\le t$, let
$T(s,t)=\sum_{u=s+1}^{t}\delta_u$ and $R(s,t)=e^{-\lambda T(s,t)}$ as in
Eq.~\ref{eq:retention}, and define
\begin{equation}
  P_{s,t}=B_{s+1}\cdots B_t,\qquad
  C^{(\kappa)}_{s,t}=R(s,t)^{\kappa}P_{s,t},
  \label{eq:influence}
\end{equation}
with $P_{t,t}=I$. The $B_u$ need not commute, so the order of the product
matters.

\begin{lemma}\label{lem:erase}
$\|B_t\|_2\le 1$ for every $t$, and hence $\|P_{s,t}\|_2\le 1$.
\end{lemma}
\begin{proof}
Since $\|k_t\|_2=1$, the matrix $k_tk_t^{\top}$ is the orthogonal projection
onto the span of $k_t$. The symmetric matrix $B_t$ therefore has eigenvalue
$1-\beta_t$ along $k_t$ and eigenvalue $1$ on its orthogonal complement.
Both lie in $[0,1]$ for $\beta_t\in[0,1]$, so $\|B_t\|_2\le 1$. The bound
on $P_{s,t}$ follows because the spectral norm is submultiplicative.
\end{proof}
\subsection{Gate, retention and half-life (P1, P2)}

\begin{proposition}\label{prop:power}
Under $E_\kappa$, every gate becomes $\alpha_t^{\kappa}$ and every share
$R(s,t)$ becomes $R(s,t)^{\kappa}$. Measured in clock time, the half-life
of the gate share is $\ln 2/(\kappa\lambda)$. If the clock is constant,
$\delta_t=\bar\delta$, its half-life in tokens is
$\ln 2/(\kappa\lambda\bar\delta)$.
\end{proposition}
\begin{proof}
By Eqs.~\ref{eq:gate} and~\ref{eq:edit}, the edited gate is
$\exp(-\kappa\lambda\delta_t)=\alpha_t^{\kappa}$. Taking the product over
$u=s+1,\dots,t$ gives
$R^{(\kappa)}(s,t)=\exp(-\kappa\lambda T(s,t))=R(s,t)^{\kappa}$. This share
equals $1/2$ exactly when $T(s,t)=\ln 2/(\kappa\lambda)$. With a constant
clock, $T(s,t)=(t-s)\bar\delta$, which gives the half-life in tokens. When
the clock varies, the number of tokens needed to reach this clock time
depends on the sequence of $\delta_t$, so the factor $\kappa$ is exact in
clock time but not, in general, in tokens.
\end{proof}

For the constant gate of Figure~\ref{fig:halflife}, $R=\alpha^{\kappa n}$
and the half-life is $\ln 2/(-\kappa\ln\alpha)$, which is 137.9, 69.0 and
34.5 tokens for $\kappa=0.5$, 1 and 2 at $\alpha=0.99$.

The gate share is not the whole story, because the erase term also removes
part of a written value. For a write $W_s\neq 0$ made at token $s$, let
$e(n)=\|W_sP_{s,s+n}\|_F/\|W_s\|_F$ be the part left by erasure after $n$
further tokens. By Proposition~\ref{prop:influence} below, the edited head
keeps $c_\kappa(n)=R(s,s+n)^{\kappa}e(n)$ of the write in Frobenius norm,
and under fixed inputs the edit does not change $e$.

\begin{proposition}\label{prop:survive}
$e(0)=1$, and $e(n)$ does not increase with $n$. Let $\kappa\ge 1$, let the
clock be constant, and let $h_\kappa$ be the first $n$ with
$c_\kappa(n)\le 1/2$. Then $(h_1-1)/\kappa<h_\kappa\le h_1$.
\end{proposition}
\begin{proof}
$P_{s,s}=I$ gives $e(0)=1$, and
$\|W_sP_{s,s+n+1}\|_F=\|W_sP_{s,s+n}B_{s+n+1}\|_F\le\|W_sP_{s,s+n}\|_F$ by
Lemma~\ref{lem:erase}. Since $R^{\kappa}\le R$, $c_\kappa\le c_1$, so
$c_\kappa(h_1)\le 1/2$ and $h_\kappa\le h_1$. For the lower bound, let
$n\le(h_1-1)/\kappa$. With a constant clock, $R(s,s+n)^{\kappa}=\alpha^{\kappa
n}\ge\alpha^{h_1-1}$, and $n\le h_1-1$ gives $e(n)\ge e(h_1-1)$. Hence
$c_\kappa(n)\ge c_1(h_1-1)>1/2$, so $h_\kappa>n$ for every such $n$.
\end{proof}

Faster forgetting therefore shortens the half-life of what survives by a
factor between 1 and about $\kappa$, not by exactly $\kappa$. In a scalar
head that writes to the same key at every token with a constant $\beta$,
$e(n)=(1-\beta)^n$ and $c_\kappa(n)=\bigl(\alpha^{\kappa}(1-\beta)\bigr)^n$:
at $\alpha=0.99$ and $\beta=1/2$, erasure dominates and the surviving share
halves after one token for every $\kappa$ in our range, whereas with keys
orthogonal to all later keys, $e\equiv 1$ and the half-life is that of the
gate share.

\subsection{Ratios, validity and composition (P3, P4)}

\begin{proposition}\label{prop:family}
For all $\kappa,\kappa'>0$: (i) the ratio $\lambda_h/\lambda_{h'}$ of any two
heads is unchanged, and under fixed inputs, $R_h(s,t)\ge R_{h'}(s,t)$ holds
after the edit exactly when it held before; (ii) every edited gate lies in $(0,1)$;
(iii) $E_\kappa\circ E_{\kappa'}=E_{\kappa\kappa'}$, $E_1$ is the identity,
and $E_{1/\kappa}$ undoes $E_\kappa$; (iv) for $\kappa\ge 1$,
$\alpha_t^{\kappa}\le\alpha_t$ for every $t$.
\end{proposition}
\begin{proof}
(i) Every rate is multiplied by the same $\kappa$, which cancels in the
ratio. Both shares become their $\kappa$-th powers, and $x\mapsto x^{\kappa}$
is increasing on $(0,1]$. Because the clocks of two heads differ, the
order of the rates alone does not fix the order of the shares. (ii) $\lambda=e^{A}>0$ and $\delta_t=\mathrm{softplus}(\cdot)>0$, so
$\kappa\lambda\delta_t>0$ and $\exp(-\kappa\lambda\delta_t)\in(0,1)$.
(iii) $(A+\ln\kappa')+\ln\kappa=A+\ln(\kappa\kappa')$, and $\ln 1=0$.
(iv) $0<\alpha_t<1$ and $\kappa\ge 1$ give $\alpha_t^{\kappa}\le\alpha_t$.
\end{proof}

These statements hold in exact arithmetic. With \texttt{A\_log} stored in
float32, two successive shifts can differ from a single combined shift by
rounding.

\subsection{Influence of past writes}

\begin{proposition}\label{prop:influence}
For every $\kappa>0$,
\begin{equation}
  S^{(\kappa)}_t=S_0\,C^{(\kappa)}_{0,t}+\sum_{s=1}^{t}W_s\,C^{(\kappa)}_{s,t},
  \label{eq:unrolled}
\end{equation}
where $C^{(\kappa)}_{s,t}=R(s,t)^{\kappa-1}C^{(1)}_{s,t}$ and
$\|C^{(\kappa)}_{s,t}\|_2\le R(s,t)^{\kappa}$. For $\kappa\ge 1$, let
$d_{s,t}=R(s,t)-R(s,t)^{\kappa}\ge 0$. Then
\begin{equation}
  \bigl\|S^{(\kappa)}_t-S^{(1)}_t\bigr\|_F
  \le\|S_0\|_F\,d_{0,t}+\sum_{s=1}^{t}\|W_s\|_F\,d_{s,t},
  \label{eq:perturb}
\end{equation}
and the read-outs $z_t=S_tq_t$ satisfy
$\|z^{(\kappa)}_t-z^{(1)}_t\|_2\le\|S^{(\kappa)}_t-S^{(1)}_t\|_F\,\|q_t\|_2$.
\end{proposition}
\begin{proof}
Let $M_u=\alpha_u^{\kappa}B_u$, so that
$S^{(\kappa)}_u=S^{(\kappa)}_{u-1}M_u+W_u$. Repeated substitution gives
\[
  S^{(\kappa)}_t=S_0M_1\cdots M_t+\sum_{s=1}^{t}W_sM_{s+1}\cdots M_t .
\]
The scalar factors $\alpha_u^{\kappa}$ commute with the matrices, so
$M_{s+1}\cdots M_t=\bigl(\prod_{u=s+1}^{t}\alpha_u^{\kappa}\bigr)P_{s,t}
=R(s,t)^{\kappa}P_{s,t}=C^{(\kappa)}_{s,t}$, which proves
Eq.~\ref{eq:unrolled}; the identity
$C^{(\kappa)}_{s,t}=R(s,t)^{\kappa-1}C^{(1)}_{s,t}$ follows from
Eq.~\ref{eq:influence}. Lemma~\ref{lem:erase} gives
$\|C^{(\kappa)}_{s,t}\|_2\le R(s,t)^{\kappa}$. For $\kappa\ge 1$ and
$R\in(0,1]$, $R^{\kappa}\le R$, so $d_{s,t}\ge 0$ and
$C^{(\kappa)}_{s,t}-C^{(1)}_{s,t}=-d_{s,t}P_{s,t}$. Subtracting the two
unrolled states and using $\|XP\|_F\le\|X\|_F\|P\|_2\le\|X\|_F$ term by term
gives Eq.~\ref{eq:perturb}. The bound on the read-out follows from
$\|Xq\|_2\le\|X\|_F\|q\|_2$.
\end{proof}

Proposition~\ref{prop:influence} makes the contraction in (P4) precise: for
$\kappa\ge 1$, the contribution of every past write, and of the initial
state, to the state before the per-head norm is bounded by a smaller
factor; after the norm the edit re-weights writes instead
(\S\ref{sec:readout}). The proposition does not order the norm of the
state itself, because writes of opposite sign can cancel less once the older
ones are weakened. In a scalar head ($d_k=d_v=1$) with $S_0=0$,
$k_1=k_2=1$, $\beta_1=\beta_2=1/2$, $v_1=2$, $v_2=-1/2$ and a constant gate
$\alpha=1/2$, we have $B_t=1/2$, $W_1=1$ and $W_2=-1/4$, so $S_1=1$ and
$S^{(\kappa)}_2=\alpha^{\kappa}/2-1/4$. The state is $0$ at $\kappa=1$ and
$-1/8$ at $\kappa=2$: faster forgetting makes it larger in magnitude.

\subsection{Age of what the state remembers}

Fix a head and a time $t$. For each source $s\in\{0,1,\dots,t\}$, where
$s=0$ stands for the initial state, let $x_s=T(s,t)$ be its age in clock time
and let $w_s\ge 0$ be fixed weights, not all zero. One natural choice is the
size of a write after erasure and before decay,
$w_s=\|W_sP_{s,t}\|_F$ (and $w_0=\|S_0P_{0,t}\|_F$); by
Proposition~\ref{prop:influence}, the edited head keeps exactly
$w_sR(s,t)^{\kappa}=w_se^{-\kappa\lambda x_s}$ of it. Normalizing these sizes
defines a distribution over ages,
\begin{equation}
  p_\kappa(s)=\frac{w_s e^{-\kappa\lambda x_s}}{Z_\kappa},\qquad
  Z_\kappa=\sum_{s}w_s e^{-\kappa\lambda x_s}.
  \label{eq:agedist}
\end{equation}
It describes the sizes of the surviving contributions, not their signed sum.

\begin{proposition}\label{prop:age}
Let $X$ be an age drawn from $p_\kappa$. For every nondecreasing function $f$,
\begin{equation}
  \frac{d}{d\kappa}\,\mathbb{E}_\kappa[f(X)]
  =-\lambda\,\mathrm{Cov}_\kappa\bigl(f(X),X\bigr)\le 0 .
\end{equation}
In particular, the mean age and the share of ages above any threshold do not
increase with $\kappa$, and
$\frac{d}{d\kappa}\mathbb{E}_\kappa[X]=-\lambda\,\mathrm{Var}_\kappa(X)$.
\end{proposition}
\begin{proof}
Differentiating Eq.~\ref{eq:agedist} gives
$\frac{d}{d\kappa}p_\kappa(s)=-\lambda\,p_\kappa(s)\bigl(x_s-\mathbb{E}_\kappa[X]\bigr)$.
Multiplying by $f(x_s)$ and summing over $s$ gives the covariance identity.
Let $X'$ be an independent draw from $p_\kappa$. Then
$2\,\mathrm{Cov}_\kappa(f(X),X)=\mathbb{E}_\kappa\bigl[(f(X)-f(X'))(X-X')\bigr]$,
and every term of this expectation is nonnegative because $f$ is
nondecreasing. Taking $f(x)=x$ gives the statement on the mean, and
$f(x)=\mathbf{1}\{x\ge u\}$ the statement on the share of ages of at least
$u$.
\end{proof}

For a constant clock and equal weights $w_s$ over an unbounded past, for
example writes to keys orthogonal to all later keys, so that nothing is
erased, $p_\kappa(n)=(1-\alpha^{\kappa})\alpha^{\kappa n}$ and the mean age is
$\alpha^{\kappa}/(1-\alpha^{\kappa})$ tokens: 99 at $\alpha=0.99$ and
$\kappa=1$, and about 49 at $\kappa=2$. Equal raw writes are not enough.
If every later write erases a share $\beta$ along the same key,
$w_s\propto(1-\beta)^{t-s}$, the ratio becomes $\alpha^{\kappa}(1-\beta)$,
and at $\alpha=0.99$ and $\beta=1/2$ the mean age is about one token for
both values of $\kappa$.

\subsection{Read-out scale}
\label{sec:readout}

\begin{proposition}\label{prop:scale}
(i) For a constant gate $\alpha\in(0,1)$, the geometric gain
$G(\alpha)=\sum_{n\ge 0}\alpha^n=1/(1-\alpha)$ satisfies
$G(\alpha^{\kappa})/G(\alpha)\to 1/\kappa$ as $\alpha\to 1$ and
$G(\alpha^{\kappa})/G(\alpha)\to 1$ as $\alpha\to 0$. The newest write's
share of the gain, $1/G(\alpha^{\kappa})=1-\alpha^{\kappa}$, exceeds its
share before the edit by the factor $(1-\alpha^{\kappa})/(1-\alpha)$, which
lies in $[1,\kappa]$ for $\kappa\ge 1$ and tends to $\kappa$ as
$\alpha\to 1$, while the share of all writes at least $m$ tokens old falls
from $\alpha^m$ to $\alpha^{\kappa m}$.
(ii) Let $N_\epsilon(z)=\gamma\odot z/\sqrt{\|z\|_2^2/d+\epsilon}$ be an RMS
norm with gain $\gamma\in\mathbb{R}^d$ and $\epsilon>0$, and let
$m=\|z\|_2^2/d$. For every $c>0$,
\begin{equation}
  N_\epsilon(cz)=\sqrt{\frac{m+\epsilon}{m+\epsilon/c^2}}\;N_\epsilon(z),
\end{equation}
and for $0<c\le 1$ the factor lies in $[c,1]$ and tends to $1$ as
$\epsilon/(c^2m)\to 0$.
\end{proposition}
\begin{proof}
(i) $G(\alpha^{\kappa})/G(\alpha)=(1-\alpha)/(1-\alpha^{\kappa})$. As
$\alpha\to 1$, l'H\^opital's rule gives the limit
$1/(\kappa\alpha^{\kappa-1})\to 1/\kappa$; as $\alpha\to 0$, numerator and
denominator both tend to $1$. For the shares, $\alpha^{\kappa}\le\alpha$
gives the lower bound $1$; Bernoulli's inequality,
$(1-x)^{\kappa}\ge 1-\kappa x$ for $x=1-\alpha$ and $\kappa\ge 1$, gives
$1-\alpha^{\kappa}\le\kappa(1-\alpha)$; and the limit is the reciprocal of
the one above. The writes at least $m$ tokens old carry
$\sum_{n\ge m}\alpha^{\kappa n}/G(\alpha^{\kappa})=\alpha^{\kappa m}$ of
the gain. (ii)
$N_\epsilon(cz)=c\,\gamma\odot z/\sqrt{c^2m+\epsilon}
=\gamma\odot z/\sqrt{m+\epsilon/c^2}$, and dividing by $N_\epsilon(z)$
gives the factor. For $c\le 1$, $\epsilon/c^2\ge\epsilon$ gives the upper
bound $1$, and $m+\epsilon\ge c^2(m+\epsilon/c^2)=c^2m+\epsilon$ gives the
lower bound $c$.
\end{proof}

Part (ii) is the sense in which a uniform rescaling of a head's read-out is
normalized away up to $\epsilon$ (\S\ref{sec:method}). The output gate
that follows the norm is computed from the layer's input, not from the
read-out, so under fixed inputs it is the same in both runs. The edit,
however, does not rescale the read-out uniformly: by
Proposition~\ref{prop:influence}, it multiplies the contribution of a write
made at $s$ by $R(s,t)^{\kappa-1}$, which depends on its age, so the
direction of the read-out changes as well.

\subsection{No fixed direction for reasoning length}

\begin{proposition}\label{prop:stopping}
Even with fixed inputs, and with each head's read-out passed through its own
RMS norm as in Qwen3.5, an edit that forgets faster can make a generation
either longer or shorter; the direction depends on how the normalized
read-out is mapped to the decision to stop.
\end{proposition}
\begin{proof}
Take a head with one-dimensional keys and two-dimensional values
($d_k=1$, $d_v=2$), $S_0=0$, $k_1=k_2=q_2=1$, $\beta_1=\beta_2=1/2$,
$v_1=(2,0)^{\top}$, $v_2=(0,2)^{\top}$ and a constant gate
$\alpha\in(0,1)$. Then $B_t=1/2$, $W_1=(1,0)^{\top}$ and
$W_2=(0,1)^{\top}$, so $S_1=(1,0)^{\top}$ and the read-out at step 2 is
$z^{(\kappa)}=S^{(\kappa)}_2q_2=(x_\kappa,1)^{\top}$ with
$x_\kappa=\alpha^{\kappa}/2$, which decreases in $\kappa$ because
$dx_\kappa/d\kappa=\tfrac12\alpha^{\kappa}\ln\alpha<0$. Apply the RMS norm
of Proposition~\ref{prop:scale} with unit gain and any $\epsilon>0$ (the argument also holds at
$\epsilon=0$). Its
first coordinate is
\begin{equation}
  \begin{gathered}
  u_\kappa=\frac{x_\kappa}{\sqrt{(x_\kappa^2+1)/2+\epsilon}},\\
  \frac{du_\kappa}{dx_\kappa}
  =\frac{1/2+\epsilon}{\bigl((x_\kappa^2+1)/2+\epsilon\bigr)^{3/2}}>0,
  \end{gathered}
\end{equation}
so $du_\kappa/d\kappa<0$: the norm removes the overall scale of the
read-out, but faster forgetting still turns its direction away from the
older write. Let generation stop after step 2 with probability
$p_\kappa=\sigma(cu_\kappa+b)$ and after step 3 otherwise, where $\sigma$ is
the logistic function and $b,c\in\mathbb{R}$; the output gate that follows
the norm depends only on the layer input, so under fixed inputs it rescales
$u_\kappa$ by a constant that can be absorbed into $c$. The expected length
is $\mathbb{E}[L_\kappa]=3-p_\kappa$, so
\begin{equation}
  \frac{d\,\mathbb{E}[L_\kappa]}{d\kappa}
  =-c\,\sigma'\bigl(cu_\kappa+b\bigr)\,\frac{du_\kappa}{d\kappa},
\end{equation}
which has the sign of $c$. Faster forgetting therefore lengthens the
generation for $c>0$ and shortens it for $c<0$.
\end{proof}

Propositions~\ref{prop:power}--\ref{prop:scale} describe one layer with its
inputs held fixed. Proposition~\ref{prop:stopping} shows that these
properties alone do not determine whether reasoning gets shorter, and in the
full network the inputs also change during generation. This is why
\S\ref{sec:experiments} measures the effect of $\kappa$ on reasoning
length rather than deriving it.

\end{document}